\documentclass[a4paper,11pt]{article}
\usepackage{tablefootnote}
\usepackage{eurosym}
\usepackage{longtable}
\usepackage{tabu}
\usepackage{geometry}
\usepackage[bottom]{footmisc}
\usepackage{tikz}
\usepackage{amsmath}
\usepackage{bbm}
\usepackage{bm}
\usepackage{amssymb}
\usepackage{amsfonts}
\usepackage{sidecap}
\usepackage{cellspace}
\usepackage{makecell}
\usepackage{amssymb}
\usepackage{breqn}
\usepackage{scalerel}
\usepackage{multicol}
\usepackage{varwidth}
\usepackage[T1]{fontenc}
\usepackage{tabularx}
\usepackage{xcolor, colortbl}
\usepackage{amsmath}
\usepackage{systeme}
\usepackage[font = small, labelfont = bf]{caption}
\usepackage{tocloft}
\usepackage{titlesec}
\usepackage{booktabs,caption}
\usepackage[flushleft]{threeparttable}
\usepackage{etoolbox}
\usepackage{hyperref}
\usepackage[document]{ragged2e}
\usepackage{tikz}
\usepackage[backend=biber, style=ieee]{biblatex}
\usepackage[utf8]{inputenc}
\usepackage[english]{babel}
\usepackage{amsthm}
\usepackage{amsthm}
\usepackage{pgfplots}
\usepackage{wrapfig}
\usepackage{algorithm}
\usepackage{algcompatible}
\usepackage{amsmath}
\usepackage[noend]{algpseudocode}
\usepackage{array}
\usepackage{multirow}
\usepackage[justification=centering]{caption}
\usepackage{amsthm}
\usepackage{siunitx,tabularx,ragged2e,booktabs,caption}
\usepackage[final]{pdfpages}
\usepackage{booktabs,caption}
\usepackage[flushleft]{threeparttable}
\usepackage{float}
\usepackage{lscape}
\usepackage{diagbox}
\usepackage{adjustbox}
\usepackage{booktabs}
\usepackage{multirow}
\usepackage{fancyhdr}
\usepackage{blindtext}
\usepackage[utf8]{inputenc}
\usepackage{graphicx}
\usepackage{mathtools}
\usepackage{gensymb}
\usepackage{amssymb}
\usepackage{amsmath}
\usepackage{latexsym}
\usepackage{amsthm}
\usepackage{eucal}
\usepackage{listings}
\usepackage{thmtools}
\usepackage{mdframed}
\usepackage[space]{grffile}

\tikzstyle{input}=[draw,circle,minimum size=20pt,inner sep=0pt]
\tikzstyle{hidden}=[draw,circle,minimum size=20pt,inner sep=0pt]
\tikzstyle{output}=[draw,circle,minimum size=20pt,inner sep=0pt]
\tikzstyle{bias}=[draw,dashed,fill=gray!50,circle,minimum size=20pt,inner sep=0pt]
\tikzstyle{layer}=[fill=gray!10]
\tikzstyle{stateTransition}=[->, thick]

\titleformat{\chapter}
 {\Large\bfseries} 
 {}                
 {0pt}             
 {\huge}           
 \titlespacing{\chapter}{0pt}{0pt}{0pt} 
\makeatletter
\patchcmd{\chapter}{\if@openright\cleardoublepage\else\clearpage\fi}{}{}{}
\makeatother
\usetikzlibrary{matrix,chains,positioning,decorations.pathreplacing,arrows}
\tikzstyle{process} = [rectangle, minimum width=3cm, minimum height=1cm, text centered, text width=3cm, draw=black, fill=orange!30]
\tikzstyle{startstop} = [rectangle, rounded corners, minimum width=3cm, minimum height=1cm,text centered, draw=black, fill=red!30]
\tikzstyle{io} = [trapezium, trapezium left angle=70, trapezium right angle=110, minimum width=3cm, minimum height=1cm, text centered, draw=black, fill=blue!30]
\tikzstyle{process} = [rectangle, minimum width=3cm, minimum height=1cm, text centered, draw=black, fill=orange!30]
\tikzstyle{decision} = [diamond, minimum width=3cm, minimum height=1cm, text centered, draw=black, fill=green!30]
\tikzstyle{arrow} = [thick,->,>=stealth]
\makeatletter
\renewcommand{\itemize}[1][]{ \beamer@ifempty{#1}{}{\def\beamer@defaultospec{#1}} \ifnum \@itemdepth >2\relax\@toodeep\else
   \advance\@itemdepth\@ne
   \beamer@computepref\@itemdepth   \usebeamerfont{itemize/enumerate \beameritemnestingprefix body}   \usebeamercolor[fg]{itemize/enumerate \beameritemnestingprefix body}   \usebeamertemplate{itemize/enumerate \beameritemnestingprefix body begin}   \list
     {\usebeamertemplate{itemize \beameritemnestingprefix item}}
     {\def\makelabel##1{         {           \hss\llap{{               \usebeamerfont*{itemize \beameritemnestingprefix item}               \usebeamercolor[fg]{itemize \beameritemnestingprefix item}##1}}         }       }     }
 \fi \beamer@cramped \justifying  \beamer@firstlineitemizeunskip}
\makeatother
\usetikzlibrary{matrix}
\theoremstyle{definition}

\makeatletter
\def\BState{\State\hskip-\ALG@thistlm}
\makeatother
\newcommand\MyBox[2]{
  \fbox{\lower0.75cm
    \vbox to 2.7cm{\vfil
      \hbox to 2.7cm{\hfil\parbox{1.4cm}{#1\\#2}\hfil}
      \vfil}  }}
\theoremstyle{plain}

\theoremstyle{definition}

\theoremstyle{remark}

\newcolumntype{Y}{>{\RaggedRight\arraybackslash}X}
\newcolumntype{T}[1]{S[table-format=#1]}  
\makeatletter
\patchcmd\thmt@mklistcmd
 {\thmt@thmname}
 {\check@optarg{\thmt@thmname}}
 {}{}
\patchcmd\thmt@mklistcmd
 {\thmt@thmname\ifx}
 {\check@optarg{\thmt@thmname}\ifx}
 {}{}
\protected
\def\check@optarg#1{ \@ifnextchar\thmtformatoptarg\@secondoftwo{#1}}
\makeatother
\declaretheoremstyle[
 spaceabove=6pt, spacebelow=6pt,
 headfont=\normalfont\bfseries,
 postheadspace=1em,
 notefont=\bfseries,
 notebraces={(}{)},
 bodyfont=\itshape
]{thmstyle}

\makeatletter
\AtBeginDocument{\renewcommand{\listofalgorithms}{\@cfttocstart \par \begingroup 
\parindent\z@ \parskip\cftparskip 
\addpenalty \@secpenalty 
\if@cfthaschapter \vspace*{\cftbeforeloftitleskip } \else 
                 \vspace {\cftbeforeloftitleskip } \fi
\@cftpagestyle 
{\interlinepenalty \@M 
 {\cftloftitlefont \listalgorithmname } {\cftafterloftitle } 
\@mkboth {\MakeUppercase \listalgorithmname }
        {\MakeUppercase \listalgorithmname }
\par \nobreak \vskip \cftafterloftitleskip \@afterheading }\let\l@algorithm\l@figure
\@starttoc {loa}\endgroup 
\@cfttocfinish }}
\makeatother
\makeatletter
\AtBeginDocument{\renewcommand{\listofalgorithms}{\@cfttocstart \par \begingroup 
\parindent\z@ \parskip\cftparskip 
\addpenalty \@secpenalty 
\if@cfthaschapter \vspace*{\cftbeforeloftitleskip } \else 
                 \vspace {\cftbeforeloftitleskip } \fi
\@cftpagestyle 
{\interlinepenalty \@M 
 {\cftloftitlefont \listalgorithmname } {\cftafterloftitle } 
\@mkboth {\MakeUppercase \listalgorithmname }
        {\MakeUppercase \listalgorithmname }
\par \nobreak \vskip \cftafterloftitleskip \@afterheading }\let\l@algorithm\l@figure
\@starttoc {loa}\endgroup 
\@cfttocfinish }}
\makeatother
\newcommand{\listequationsname}{List of Equations}
\newlistof{myequations}{equ}{\listequationsname}

\input{tcilatex}

\begin{document}

\title{Ensemble Predictors for Deep Neural Networks}

\author{\textsc{Hector Calvo-Pardo}\footnote{University of Southampton, Centre for Economic and Policy Research (CEPR), Centre for Population Change (CPC), and Institut Louis Bachelier (ILB).} \hspace{3ex} \textsc{Tullio Mancini}\footnote{University of Southampton. Corresponding address: Department of Economics, University of Southampton. Highfield Campus, SO17 1BJ, Southampton. E-mail: T.Mancini@soton.ac.uk. Tullio Mancini acknowledges financial support from The University of Southampton Presidential Scholarship.} \hspace{3ex} \textsc{Jose Olmo}\footnote{University of Southampton and Universidad de Zaragoza. Jose Olmo acknowledges financial support from Fundaci\'on Aragonesa para la Investigaci\'on y el Desarrollo (ARAID).}}

\maketitle

\justify

\begin{abstract}
The aim of this paper is to propose a novel prediction model based on an ensemble of deep neural networks. To do this, we adapt the extremely randomized trees method originally developed for random forests. The extra-randomness introduced in the ensemble reduces the variance of the predictions and yields gains in out-of-sample accuracy. As a byproduct, we are able to compute the uncertainty about our model predictions and construct interval forecasts. An extensive Monte Carlo simulation exercise shows the good performance of this novel prediction method in terms of mean square prediction error and the accuracy of the prediction intervals in terms of out-of-sample coverage probabilities. This approach is superior to state-of-the-art methods extant in the literature such as the widely used Monte Carlo dropout and bootstrap procedures. The out-of-sample accuracy of the novel algorithm is further evaluated using experimental settings already adopted in the literature.

\vspace{7mm}

\noindent \textbf{Keywords:} Neural networks, ensemble methods, prediction interval, uncertainty quantification, dropout

\end{abstract}


\newpage
\pagestyle{plain} 

\baselineskip19.5pt


\newpage

\section[Introduction]{Introduction}\label{sec1} 

A popular and fruitful strategy in the prediction literature is model averaging. Steel (2020)  distinguishes two main categories: Bayesian model averaging (see Leamer, 1978), where the model index is treated as unknown (and thus a prior is specified on both model and model parameters); and frequentist model averaging (see Wang et al., 2009; Dormann et al., 2018), where the predictions of a battery of different prediction models are ensembled. There is a well established theoretical and empirical literature analyzing the predictive advantages of both Bayesian and frequentist approaches. 

When focusing on Bayesian model averaging, Min and Zellner (1993) show that the expected squared errors are minimized by Bayesian ensembles as long as the model underlying the data generating process is included in the model space. Fernandez et al. (2001) explain how Bayesian ensembles improve in terms of predictability over single models when dealing with growth data. More generally, the empirical study conducted by Raftery et al. (1997) highlights how Bayesian model averaging improves over single model predictions. As explained in Steel (2020), one of the main advantages of Bayesian ensembling is the possibility of integrating the prior structure analytically; nonetheless, a vast model space may constitute a computational challenge due to the impossibility of a complete model enumeration (this problem is often solved with Markov chain Monte Carlo (MCMC) methods). 

Steel (2020) highlights that MCMC methods are not implementable for frequentist model averaging approaches (no estimation of the model probabilities), ultimately limiting the possibility of applying frequentist approaches to a large number of models. As a result, the literature focusing on frequentist ensembles tries to propose methods directed to reducing the model space (see, for example, Claeskens et al., 2006; Zhang et al., 2016; and Zhang et al., 2013). Finally, when analyzing frequentist model averaging, it is essential to mention the research conducted by Stock and Watson (2004). After examining different weighting schemes found in the literature, these authors show how ''optimally'' estimated weights perform worse than equal weights in terms of out-of-sample mean squared error. Smith and Wallis (2009) explain that this ''forecast combination puzzle'' can be explained starting from the double estimation uncertainty associated with estimating the ''optimal'' weights.

Similarly, model averaging in machine learning (e.g., boosting, bagging, random forest, and extremely randomized trees) aims to construct a predictive model by combining ''weak'' learners to obtain a strong learner. As opposed to the aforementioned literature, model averaging in machine learning does not allow the estimation of the uncertainty on the parameter estimates and the identification of the model structure. Instead, it is mainly focused on point prediction/forecasting and associated uncertainty estimation. 

Neural networks are increasingly popular prediction models in the machine learning literature. These models are widely used in prediction tasks due to their unrivaled performance and flexibility in modeling complex unknown functions of the data. Although these methods provide accurate predictions, the development of tools to estimate the uncertainty around their predictions is still in its infancy. As explained in H\"{u}llermeier and Waegeman (2020) and Pearce et al. (2018), out-of-sample pointwise accuracy is not enough\footnote{A trustworthy representation of uncertainty can be considered pivotal when machine learning techniques are applied to medicine (Yang et al., 2009; Lambrou et al., 2011), or to anomaly detection, optimal resource allocation and budget planning (Zhu and Laptev, 2017), or cyber-physical systems (Varshney and Alemzadeh, 2017) defined as surgical robots, self-driving cars and the smart grid.}. The predictions of deep neural network (DNN) models need to be supported by measures of uncertainty in order to provide satisfactory answers for prediction in high-dimensional regression models, pattern recognition, biomedical diagnosis, and others (see Schmidhuber (2015) and LeCun et al. (2015) for overviews of the topic). 

The present paper focuses on a machine learning approach for model prediction. We propose an ensemble of neural network models with the aim of improving the accuracy of existing model predictions from individual neural networks. A second main contribution of the present study is to assess the uncertainy about the predictions of these ensembles of neural network models and construct interval forecasts. Our novel approach extends the Extra-trees algorithm (Geurts et al., 2006) to ensembles of deep neural networks using a fixed Bernoulli mask. To do this, we estimate $T$ different sub-networks with randomized architectures (each network will have different layer-specific widths) that are independently trained on the same dataset. Thus, the fixed Bernoulli mask introduces an additional randomization scheme to the prediction obtained from the ensemble of neural networks that ensures independence between the components of the ensemble reducing, in turn, the variance associated to the prediction and yielding accurate prediction intervals. Additionally, based on the findings of Lee et al. (2015) and Lakshminarayanan et al. (2017), the novel procedure is expected to outperform bootstrap based approaches in terms not only of estimation accuracy but also of uncertainty estimation. This is confirmed in our simulation experiments.

The competitors of our ensemble prediction model are found in the machine learning literature. In particular, we consider Monte Carlo dropout and bootstrap procedures as the benchmark models to beat in out-of-sample prediction exercises. Monte Carlo dropout approximates the predictive distribution of a target variable by fitting a deep or shallow network with dropout both at train and test time. Conversely, both extra-neural network and bootstrap based approaches approximate the target predictive distribution via ensemble methods. When comparing classical bootstrap approaches to the extra-neural network approach proposed in this paper, we notice that (i) both methods guarantee conditional randomness of the predicted outputs, the extra-neural network method does it through the Bernoulli random variables with probability $p$ and random weight initialization, whereas the bootstrap does it through the nonparametric data resampling and random weight initialization; (ii) by performing data resampling, the naive (nonparametric) bootstrap approach requires the assumption that observations are independent and identically distributed (\textit{i.i.d}). Importantly, each single model is trained with only $63\%$ unique observations of the original sample due to resampling with replacement; (iii) by randomizing the neural network structures, the extra-neural network approach increases the diversity (see Zhou (2012) for an analysis of diversity and ensemble methods) among the individual learners; and (iv) the extra-neural network will benefit from the generalization gains associated with dropout (one can think of the dropout approach of Srivastava et al. (2014) as an ensemble of sub-networks trained for one gradient step).

To analyze the out-of-sample performance and the empirical coverage probabilities of the proposed methodologies, we carry out an extensive Monte Carlo exercise that evaluates the Monte Carlo dropout, the bootstrap approach, and extra-neural network for both deep and shallow neural networks given different dropout rates and data generating processes. The simulation results show that all three procedures return prediction intervals approximately equal to the theoretical ones for nominal values equal to $0.01$ and $0.05$; for prediction intervals constructed at $0.10$ significance level, the extra-neural network is shown to outperform both Monte Carlo dropout and bootstrap. Additionally, the simulation findings show that the extra-neural network approach returns prediction intervals with correct empirical coverage rates for different dropout rates (within a reasonable range) as opposed to the MC dropout that returns correct prediction intervals for specific values of the dropout rate. The findings not only show the robustness of the extra-neural network to the choice of the dropout rate, but they also complete the results of Levasseur et al. (2017) by showing that the Monte Carlo dropout returns correct prediction intervals when the dropout rate that yields the highest out-of-sample accuracy is adopted.

Finally, the novel methodology is also evaluated on real world datasets. In order to allow for comparability with other approaches found in the literature, the experimental settings of Hern\'{a}ndez-Lobato and Adams (2015) are adopted. The empirical results show that extra-neural network methods outperform other state-of-the-art approaches used in the literature. These results complete the conclusions drawn from the Monte Carlo simulation by showing the generalization of the extra-neural network methodology when applied to large dimensional datasets. 

The current study is related to a recent literature on prediction intervals for neural networks. A pioneering contribution is provided by Hwang and Ding (1997) that construct asymptotically valid prediction intervals for neural networks. Yet, being their research focused only on single layer feedforward neural networks with sigmoidal activation function, it does not find applicability in some widely adopted neural network structures (i.e., convolutional neural networks, recurrent neural networks, and deep feedforward neural networks with the ReLu activation function). Importantly, this early work on prediction intervals on neural networks does not incorporate recent advances in machine learning prediction such as deep learning models and Monte Carlo dropout. This technique - proposed by Srivastava et al. (2014) - ensures better generalization for neural networks by forcing the hidden nodes not to co-adapt with the neighboring nodes.

Levasseur et al. (2017) notice that one of the main obstacles for assessing uncertainty around the outputs of neural network models is the fact that the weights characterizing the predictions are usually fixed, implying that the output is deterministic. In contrast, Bayesian neural networks (Denker and LeCun, 1991) - instead of defining deterministic weights - allow the networks' weights to be defined by a given probability distribution and can capture the posterior distribution of the output, providing a probabilistic measure of uncertainty around the model predictions. Being the approximation of the posterior distribution a difficult task, the literature focusing on deep Bayesian neural networks has proposed different alternatives for the estimation of such distribution. These alternatives center around the Bayesian interpretation of dropout methods to estimate the uncertainty in the model predictions. A noteworthy example is Gal and Ghahramani (2016a); these authors develop a Monte Carlo dropout to model both parameter and data uncertainty by fitting a deep neural network with dropout implemented not only at training but also during test phase. During test time, each forward pass is multiplied by a random variable to generate a random sample of the approximated posterior distribution. Levasseur et al. (2017) analyze the coverage probability of the procedure proposed by Gal and Ghahramani (2016a) and conclude that the construction of prediction intervals with correct empirical coverage probabilities is highly dependent on the adequate tuning of the dropout rate. 

Applying dropout during the test phase can also be regarded as an approach to estimate the uncertainty around the predicted outputs from deep neural networks that works outside the Bayesian framework (for example, Cortes-Ciriano and Bender, 2019). However, any suitable method that aims at constructing valid prediction intervals based on the Monte Carlo dropout must also incorporate the uncertainty due to noise in the data. It is based on this final aspect that Kendall and Gal (2017), Serpell et al. (2019), and Zhu and Laptev (2017) propose novel methodologies for the correct estimation of the prediction uncertainty for both shallow and deep networks. To do so, Kendall and Gal (2017) propose a new loss function that allows estimating the aleatoric uncertainty from the input data; Serpell et al. (2019) couple the stochastic forward passes of the Monte Carlo dropout with the Mean Variance Estimation\footnote{The Mean Variance Estimation method - introduced by Nix and Weigend (1994) - involves fitting a neural network with two output nodes capturing the mean and the variance, respectively, of a Normal distribution.}; and Zhu and Laptev (2017) propose to estimate the data uncertainty with a consistent estimator in a hold out set. 

Another branch of the literature has been focusing on adopting bootstrap based approaches for the estimation of the prediction intervals of neural networks (see for example, Carney et al., 1999;  and Errouissi et al., 2015). Bootstrap procedures have become increasingly popular, despite their computational requirements, as they provide a reliable solution to obtain the predictive distribution of the output variable in both shallow and deep neural networks. Recent advances in the neural network literature (Pearce et al., 2018; Lee et al., 2015; and Lakshminarayanan et al., 2017) have also shown how parameter resampling without data resampling can improve over standard bootstrap approaches not only in terms of out-of-sample accuracy but also in terms of prediction uncertainty estimation. 

The rest of the paper is organized as follows: Section \ref{ensemblesection} introduces the ensemble prediction model in a deep neural network and discusses the different sources of uncertainty. Section \ref{prediction} reviews extant methodologies to construct prediction intervals that can be applied to DNNs. Section \ref{predictionnovel} introduces a novel methodology to construct prediction intervals based on an adaptation of Extra-trees for random forests. Section \ref{montecarlo} presents the simulation setup including linear and nonlinear models along with the choice of parameters and hyperparameters for the implementation of neural network methods. Section \ref{empiricalanalysislabel} discusses the results of the empirical study. Section \ref{conclusionSection} concludes. A mathematical appendix contains a brief note discussing random weight initialization and uncertainty for extra-neural networks. 

\section[Ensemble predictors for DNN models]{Ensemble predictors for DNN models}\label{ensemblesection}

We propose the following additive model for predicting the output variable $y_i$, for $i=1,\ldots,n$:
\begin{equation} \label{model1}
y_{i} = f(\mathbf{x}_{i}) + \epsilon_{i},
\end{equation}
with $f(\mathbf{x}_{i})$ a real-valued function used to predict the outcome variable using a set of covariates $\mathbf{x}_i$. The choice of the functional form $f(\mathbf{x}_{i})$ depends on the loss function penalizing the difference between the outcome variable and the prediction. For example, it is well known that if the loss function is quadratic then the best predictive model is $f(\mathbf{x}_{i}) = E[y_i \ | \ \mathbf{x}_i]$. The error term $\epsilon$ defines the noise in the output variable that cannot be explained by the covariates $\mathbf{x}$ and satisfies the conditional independence assumption $E[\epsilon_i \ | \ \mathbf{x}_i]=0$. 

In this paper, we consider $f(\mathbf{x}_{i})$ to be modeled by a ReLu deep neural network. For any two natural numbers $d, \ n_{1} \in \mathbb{N}$, which are called input and output dimension respectively, a $\mathbb{R}^{d} \rightarrow \mathbb{R}^{n_{1}}$ ReLu DNN is given by specifying a natural number $N \in \mathbb{N}$, a sequence of $N$ natural numbers $Z_{1}, Z_{2}, \cdots, Z_{N}$, and a set of $N+1$ affine transformations $\mathbf{T}_{1}: \mathbb{R}^{d} \rightarrow \mathbb{R}^{Z_{1}}, \mathbf{T}_{i}: \mathbb{R}^{Z_{i-1}} \rightarrow \mathbb{R}^{Z_{i}}$, for $i = 2, \cdots, N$, and $\mathbf{T}_{N+1}: \mathbb{R}^{Z_{N}} \rightarrow \mathbb{R}^{n_{1}}$. Such a ReLu DNN is called a $(N+1)$-layer ReLu DNN, and is said to have $N$ hidden layers. The function $f: \mathbb{R}^{d} \rightarrow \mathbb{R}^{n_{1}}$ is the output of this ReLu DNN that is constructed as
\begin{equation}\label{eq000}
f(\mathbf{x}_{i};\bm{\omega}) = \mathbf{T}_{N+1} \circ \bm{\theta} \circ \mathbf{T}_{N} \circ \cdots \circ \mathbf{T}_{2} \circ \bm{\theta} \circ \mathbf{T}_{1},
\end{equation}

\noindent with $\mathbf{T}_{n} = \mathbf{W}^{n}\mathbf{h}_{n-1} + \mathbf{b}_{n}$, where - for $N = 1$ - $\mathbf{W}^{n} \in \mathbb{R}^{Z_{1} \times d}$; $\mathbf{h}_{0} \equiv \mathbf{x}$, with $\mathbf{x} \in \mathbb{R}^{d \times 1}$ the input layer, and $\mathbf{b}_{n} \in \mathbb{R}^{Z_{1}}$ is an intercept or bias vector. For $N \neq 1$, $\mathbf{W}^{n} \in \mathbb{R}^{Z_{n} \times Z_{n-1}}$ is a matrix with the deterministic weights determining the transmission of information across layers; $\mathbf{h}_{n-1} \in \mathbb{R}^{Z_{n-1}}$ is a vector defined as $\mathbf{h}_{n-1} = \bm{\theta}(\mathbf{T}_{n-1})$, and $\mathbf{b}_{n} \in \mathbb{R}^{Z_{n}}$. The function $\bm{\theta}$ is a ReLu activation function defined as $\bm{\theta}(\mathbf{T}_{n-1}) = \max\{0, \mathbf{T}_{n-1}\}$ and $\bm{\omega}=(\bm{W}^n,\mathbf{b}_n)$ collects the set of estimable features of the model. The \textit{depth} of a ReLu DNN is defined as $N+1$. The width of the $n^{th}$ hidden layer is $Z_{n}$, and the \textit{width} of a ReLu DNN is $\max\{Z_{1}, \cdots, Z_{N}\}$. The \textit{size} of the ReLu DNN is $Z_{\text{tot}} = Z_{1} + Z_{2} + \cdots + Z_{N}$. The number of active weights (different from zero) - in a fully connected ReLu DNN - of the $n^{th}$ hidden layer is $w_{n} = (Z_{n} \times Z_{n-1}) + Z_{n}$. The \textit{number of active weights} in a fully connected ReLu DNN is $w_{1} + w_{2} + \cdots + w_{N}$. Under these premises, universal approximation theorems developed for ReLu DNN models (Lu et al., 2017) guarantee that $f(\mathbf{x}_{i};\bm{\omega})$ approximates the true function $f(\mathbf{x}_{i})$ in \eqref{model1} arbitrarily well. See also Cybenko (1989), Leshno et al. (1993), Hornik (1991), Lu et al. (2017), and Mei et al. (2018) for universal approximation theorems in similar contexts. 

We should note the presence of an approximation error due to replacing $f(\mathbf{x}_{i})$ by $f(\mathbf{x}_{i};\bm{\omega})$ in model \eqref{model1}, where $f(\mathbf{x}_{i};\bm{\omega})$ denotes a feasible version of the DNN model that can be estimated from the data.\footnote{The feasibility of the model entails that it is defined by a truncation of the true ReLu DNN model that approximates arbitrarily well the unknown function $f(\mathbf{x}_i)$.} The model that we consider in practice is
\begin{equation} \label{model1b0}
y_{i} = f(\mathbf{x}_{i};\bm{\omega}) + u_{i},
\end{equation}
where $u_i = \varepsilon_i + f(\mathbf{x}_{i}) - f(\mathbf{x}_{i};\bm{\omega})$. In the related literature the effect of the approximation error is usually neglected, see Pearce et al. (2018) and Heskes (1997). In practice, we estimate model \eqref{model1b0} using a training sample to obtain parameter estimates $\bm{\widehat{\omega}}$, such that the relevant empirical model is
\begin{equation} \label{model1b}
y_{i} = f(\mathbf{x}_{i};\widehat{\bm{\omega}}) + e_{i},
\end{equation}
with $f(\mathbf{x}_{i};\widehat{\bm{\omega}})$ a function that is estimated from the data and $\widehat{\bm{\omega}}$ the parameter estimates of the matrices of weights $\bf{W}^n$ and bias parameters $\bf{b}_n$ defining the DNN; $e_i$ is the residual of the model. Using expressions \eqref{model1} to \eqref{model1b}, the error term in \eqref{model1} can be decomposed as
\begin{equation}
\epsilon_i =  \underbrace{f(\mathbf{x}_i;\widehat{\bm{\omega}})  - f(\mathbf{x}_i;\bm{\omega})}_{\text{estimation error}} + \underbrace{f(\mathbf{x}_i;\bm{\omega}) - f(\mathbf{x}_i)}_{\text{bias effect}} + \underbrace{e_i}_{\text{aleatoric error}}
\end{equation}
such that the conditional variance of the output variable given the set of covariates $\mathbf{x}$, denoted as $\sigma_{\epsilon}^2$, satisfies that $\sigma_{\epsilon}^2 = \sigma_{\widehat{\bm{\omega}}}^2(\mathbf{x}_i) + \sigma_{e}^2$, with $\sigma_{\widehat{\bm{\omega}}}^2(\mathbf{x}_i)$ the epistemic uncertainty due to the estimation of the model parameters and hyperparameters (estimation effect) and $\sigma_{e}^2$ the variance due to the aleatoric error.  The bias term does not have an effect on the variance of the predictor but introduces an error in the model forecast. More formally, $E[f(\mathbf{x}_{i};\bm{\omega})] = f(\mathbf{x}_i) + \mu_i$, with $\mu_i$ a constant that captures the approximation error (bias) due to using a truncation of the \textit{asymptotic} true ReLu DNN model. In this paper we concentrate on estimating the uncertainty around the predictions, given by $\sigma_{\epsilon}^2$, however, when possible, we will also discuss the bias effect due to the approximation of the ReLu DNN model.

The distinction between epistemic and aleatoric uncertainty is extremely relevant when DNNs are considered. It has been shown that deep models, notwithstanding the high confidence in their predictions, fail on specific instances due to parameter uncertainty (see H\"{u}llermeier and Waegeman, 2020). Additionally, deep learning models are subject to drastic changes in their performance when minor changes to the dataset are engineered (well known problem of \textit{adversarial examples} in Papernot et al., 2017) implying variability in the parameter estimates. For this reason, the literature focusing on deep learning and uncertainty quantification propose algorithms that allow capturing all sources of uncertainty (see Zhu and Laptev, 2017; H\"{u}llermeier and Waegeman, 2020; Senge et al., 2014; Kull and Flach, 2014; and Varshney and Alemzadeh, 2017). 

Our main objective is to study the performance of ensembles of deep neural network models and propose a novel approach that is inspired in the extremely randomized trees method originally developed for random forests. An ensemble of predictors in our context is given by 
\begin{equation}\label{ensem}
\bar{y}(\mathbf{x}_i) = \frac{1}{T}\sum_{t=1}^{T} f_t(\mathbf{x}_i;\bm{\widehat{\omega}}^{(t)}), \ \text{for} \ i=1,\ldots,n,
\end{equation}
where $f_t(\mathbf{x}_i;\bm{\widehat{\omega}}^{(t)})$ denotes a set of $T$ different prediction models based on deep neural network models; $\bm{\widehat{\omega}}^{(t)}$ denotes the estimates of the DNN model parameters and hyperparameters. Second, we quantify the uncertainty about the predictions of the ensemble model and construct interval forecasts for the model predictions.

\section{Prediction intervals for DNN models} \label{prediction}

The prediction intervals for the output of a ReLu DNN are derived from its predictive distribution. This distribution can be approximated asymptotically using a Normal distribution; by resampling methods using bootstrap procedures; and by simulation methods using Monte Carlo dropout. In this section we review the prediction intervals obtained from these procedures.  

\subsection{Asymptotic prediction intervals (Delta Method)}

In a neural network setting we estimate the predictive variance $\sigma_{\epsilon}^2$ using the test sample, of size $n$, such that $\widehat{\sigma}_{\epsilon}^2 = \widehat{\sigma}_{\widehat{\bm{\omega}}}^2(\mathbf{x}_i) + \widehat{\sigma}_{e}^2$. Under the assumption of homoscedasticity of the error term over the test sample, we can estimate consistently the aleatoric uncertainty such that $\widehat{\sigma}_{e}^2 = \frac{1}{n}  \sum_{i=1}^{n} (y_{i} - f(\mathbf{x}_{i};\widehat{\bm{\omega}}))^2$. However, estimating the variance due to parameter estimation is cumbersome unless the specific form of the function $f(\mathbf{x}_{i};\bm{\omega})$ is known to the modeler. {Under this stringent assumption, the only uncertainty in the proposed model specification is in the choice of the model parameters $\bm{\omega}$ and hyperparameters. In this case the literature proposes the delta method to approximate the estimated function $f(\mathbf{x}_{i};\widehat{\bm{\omega}})$ under a first order Taylor expansion around the true parameter vector $\bm{\omega}$. More specifically, given a data point $\mathbf{x}_{i}$, and assuming that the number of observations $M$ is sufficiently large to ensure that $\widehat{\bm{\omega}}$ is a local approximation of the true parameter vector $\bm{\omega}$, Ungar et al. (1996) show that it is possible to linearize the neural network around the data point as: 

\begin{equation}
f(\mathbf{x}_{i};\widehat{\bm{\omega}}) = f(\mathbf{x}_{i};\bm{\omega}) + \mathbf{f}^{\intercal}_{\bm{\omega}i}(\widehat{\bm{\omega}} - \bm{\omega}) + o_P(|\widehat{\bm{\omega}} - \bm{\omega}|),
\end{equation}

\vspace{2mm}
\noindent
with $\mathbf{f}^{\intercal}_{\bm{\omega}i}$ a vector with entries $\partial f(\mathbf{x}_{i};\bm{\omega})/\partial \omega_{r}$, with $r$ the number of parameter in $\bm{\omega}$, defined as (see also De vieaux et al., 1998): 

\begin{equation}
\mathbf{f}^{\intercal}_{\bm{\omega}i}= \left [ \frac{\partial f(\mathbf{x}_{i};\bm{\omega})}{\partial \omega_{1}}, \frac{\partial f(\mathbf{x}_{i};\bm{\omega})}{\partial \omega_{2}}, \cdots, \frac{\partial f(\mathbf{x}_{i};\bm{\omega})}{\partial \omega_{r}} \right ] = \nabla_{\bm{\omega}}f(\mathbf{x}_{i};\bm{\omega})
\end{equation}

\vspace{3mm}
Following Seber and Wild (1989), the literature focusing on the delta method (see Hwang and Ding, 1997; Ungar et al., 1996; De vieaux et al., 1998) propose the following estimator of the asymptotic variance of $f(\mathbf{x}_{i};\widehat{\bm{\omega}})$ evaluated at the true parameter vector $\bm{\omega}_0$: 

\begin{equation}\label{deltaUncertainty}
\widehat{\sigma}_{\widehat{\bm{\omega}}}^2(\mathbf{x}_i) \approx \widehat{\sigma}_{e}^2[ \mathbf{f}^{\intercal}_{0i}(\mathbf{J}_{0i}^{\intercal}\mathbf{J}_{0i})^{-1}\mathbf{f}_{0i}],
\end{equation}
with $\mathbf{J}_{0i}$ the Jacobian matrix evaluated at $\bm{\omega}_0$. This is defined as

\begin{equation}
\mathbf{J}_{0i} = \left [ \frac{\partial f(\mathbf{x}_i; \bm{\omega})}{\partial \bm{\omega}} \right ]_{\bm{\omega}=\bm{\omega}_0}.
\end{equation}
Therefore, using the delta method, the corresponding asymptotic predictive variance of $y_i$ is estimated as $\widehat{\sigma}_{\epsilon}^2 = \widehat{\sigma}_{e}^2 (1 + S(\widehat{\bm{\omega}}))$, with  $S(\widehat{\bm{\omega}}) = \mathbf{f}^{\intercal}_{\widehat{\bm{\omega}}i}(\mathbf{J}_{\widehat{\bm{\omega}}i}^{\intercal}\mathbf{J}_{\widehat{\bm{\omega}}i})^{-1}\mathbf{f}_{\widehat{\bm{\omega}}i} $ and under the central limit theorem, we obtain the following asymptotic prediction interval for $y_i$: 

\begin{equation} \label{pred1}
f(\mathbf{x}_{i};\bm{\widehat{\omega}}) \pm z_{1-\alpha/2} \widehat{\sigma}_{e}\sqrt{1 + S(\widehat{\bm{\omega}})},
\end{equation}
with $z_{1-\alpha/2}$ the relevant critical value from the standard Normal distribution at an $\alpha$ significance level. 

Hwang and Ding (1997) showed that, regardless the not identifiability of the weights in a neural network, the prediction interval in \eqref{pred1} is asymptotically valid when the feedforward neural network is trained to convergence. Despite providing asymptotically valid prediction intervals, the delta method is not widely adopted by the literature focusing on uncertainty quantification and deep learning due to problems associated with the computation of the Jacobian matrix. In particular, due to the high number of parameters in $\bm{\omega}$, the complex calculation of $\mathbf{J}$ is prone to error (Tibshirani, 1996); additionally, the near singularities in the model due to overfitting (Tibshirani, 1996) or due to the small sample size (De vieaux et al., 1998) make the computation of the gradient $\mathbf{J}$ unreliable or unfeasible. 

Thus, the literature has been focusing on bootstrapping techniques for the construction of prediction intervals for neural networks. In fact, as also highlighted by Tibshirani (1996), bootstrapping prediction intervals provide a feasible alternative that does not suffer from the matrix inversion problem and does not depend on the existence of derivatives. 

\subsection[Bootstrap predictive distribution]{Bootstrap predictive distribution}\label{bootsubsection}
An alternative approach to asymptotic prediction intervals is to construct a finite-sample approximation of the prediction interval. Bootstrap procedures provide a reliable solution to obtain predictive intervals of the output variable. We proceed to explain how bootstrap works in a DNN context. The literature has developed many different forms of bootstrapping methods. One of its simplest and most popular forms is the percentile or naive bootstrap proposed by Efron (1979). Under this method observations are drawn from an independent and identically distributed sample with replacement and each observation has the same probability of being extracted. 

Let $\{\mathbf{x}_i\}_{i=1}^{M}$ be a sample of $M$ observations of the set of covariates, with $\mathbf{x}_i \in \mathbb{R}^{d}$ and $M$ the length of the train sample. Let $\{\mathbf{y}_{i}\}_{i=1}^{M} \in \mathbb{R}$ be the output variable, and define $\mathbf{x}_i^{\llcorner}=(\mathbf{x}_i,y_i) \in \mathbb{R}^{d+1}$. Applying the naive bootstrap proposed by Efron (1979) to this multivariate dataset, we generate the bootstrapped dataset $\bm{x}^{\llcorner,\star} = \{\bm{x}_{i}^{\llcorner,\star}\}_{i=1}^{M} = \{\bm{x}_{i}^{\star}, y_i^{\star}\}_{i=1}^{M}$ by sampling with replacement from the original dataset $\mathbf{x}^{\llcorner}$. By repeating this procedure $T$ times, it is possible to obtain $T$ bootstrapped samples defined as $\{\mathbf{x}^{\llcorner,\star (t)}\}_{t=1}^{T}$. Each bootstrap sample is fitted to a single neural network to obtain an empirical distribution of bootstrap predictions $f(\mathbf{x}^{\star (t)};\bm{\widehat{\omega}}^{\star (t)})$, with $\bm{\widehat{\omega}}^{\star (t)} = \{\mathbf{W}^{1, \star(t)},...,\mathbf{W}^{N,\star(t)}, b_1^{\star(t)},\ldots, b_N^{\star(t)}\}$, for $t=1,\ldots,T$. In this context, a suitable bootstrap prediction interval for $y_i$ at an $\alpha$ significance level is $\left[\widehat{q}_{\alpha/2}, \widehat{q}_{1-\alpha/2} \right]$, with $\widehat{q}_{\alpha}$ the empirical $\alpha-$quantile obtained from the bootstrap distribution of $f(\mathbf{x}_{i};\bm{\widehat{\omega}}^{\star (t)})$, for $t=1,\ldots,T$. 

Alternatively,  under the assumption that the error $\epsilon$ is normally distributed, we can refine the empirical predictive interval by using the critical value from the Normal distribution. A suitable prediction interval for $\mathbf{x}_{i}$ from the test sample, with $i=1,\ldots,n$, is
\begin{equation} \label{pred2b}
f(\mathbf{x}_{i};\bm{\widehat{\omega}}) \pm z_{1-\alpha/2} \widehat{\sigma}_{\epsilon}^{\star},
\end{equation}
with $f(\mathbf{x}_{i};\bm{\widehat{\bm{\omega}}})$ the pointwise prediction of model \eqref{model1b} and $\widehat{\sigma}_{\epsilon}^{\star2} = \widehat{\sigma}_{\widehat{\bm{\omega}}}^{\star2}(\mathbf{x}_i) + \widehat{\sigma}_{e}^{2}$. Under homoscedasticity of the error term $\epsilon_i$, the aleatoric uncertainty $\sigma_{e}^2$ is estimated from the test sample as $\widehat{\sigma}_{e}^{2} = \frac{1}{n} \sum_{i=1}^{n} \left(y_{i} - f(\mathbf{x}_{i};\widehat{\bm{\omega}}) \right)^2$, with $\widehat{\bm{\omega}}$ the set of parameter estimates obtained from the original sample $\{\mathbf{x}_i^{\llcorner}\}_{i=1}^{M}$. The epistemic uncertainty is estimated from the bootstrap samples as $\widehat{\sigma}_{\widehat{\bm{\omega}}}^{\star2}(\mathbf{x}_i) = \frac{1}{T} \sum_{t=1}^{T} [f(\mathbf{x}_{i};\bm{\widehat{\omega}}^{\star (t)}) - \bar{f}(\mathbf{x}_i)]^2$, with 
\begin{equation} \label{bootave}
\bar{f}(\mathbf{x}_{i}) = \frac{1}{T}\sum_{t=1}^{T} f(\mathbf{x}_{i};\bm{\widehat{\omega}}^{* (t)}).
\end{equation} 

Unlike for the delta method, the use of bootstrap methods allows us to ameliorate the effect of the bias in the prediction of the ReLu DNN model. The bias in model \eqref{model1b} is defined as $E[f(\mathbf{x}_i; \bm{\omega})] - f(\mathbf{x}_i)$. Therefore, a suitable estimator of this quantity is $\bar{f}(\mathbf{x}_{i}) - f(\mathbf{x}_{i};\bm{\widehat{\omega}})$, with $\bar{f}(\mathbf{x}_{i})$ defined in \eqref{bootave}, such that the above prediction interval can be refined as

\begin{equation} \label{pred2c}
f(\mathbf{x}_{i};\bm{\widehat{\omega}}) - \underbrace{(\bar{f}(\mathbf{x}_{i}) - f(\mathbf{x}_{i};\bm{\widehat{\omega}}))}_{\text{bias correction}} \pm z_{1-\alpha/2} \widehat{\sigma}_{\epsilon}^{\star} = (2 f(\mathbf{x}_{i};\bm{\widehat{\omega}}) - \bar{f}(\mathbf{x}_{i})) \pm z_{1-\alpha/2} \widehat{\sigma}_{\epsilon}^{\star}.
\end{equation}

\noindent This bootstrap prediction interval can be further refined by exploiting the average prediction in \eqref{bootave}. In this case the variance of the predictor is $\overline{\sigma}_{\widehat{\bm{\omega}}}^{\star2}(\mathbf{x}_i) = \frac{1}{T} \widehat{\sigma}_{\widehat{\bm{\omega}}}^{\star2}(\mathbf{x}_i)$ and the relevant prediction interval is obtained from substituting $f(\mathbf{x}_{i};\bm{\widehat{\omega}})$ in \eqref{pred2c} with the average prediction $\bar{f}(\mathbf{x}_{i})$, such that

\begin{equation} \label{pred2d}
\bar{f}(\mathbf{x}_{i}) \pm z_{1-\alpha/2} \widehat{\sigma}_{\epsilon}^{\star},
\end{equation}
with $\widehat{\sigma}_{\epsilon}^{\star2} = \overline{\sigma}_{\widehat{\bm{\omega}}}^{\star2}(\mathbf{x}_i) + \overline{\sigma}_{e}^{2}$, where $\overline{\sigma}_{e}^{2} = \frac{1}{n} \sum_{i=1}^{n} \left(y_{i} - \bar{f}(\mathbf{x}_{i}) \right)^2$. This expression assumes that the covariance between the predictions from the different bootstrap samples is zero. Interestingly, in this case the bias correction is not necessary unless $T$ is small. This is so because the bias term for the average predictor is negligible and given by $\frac{1}{T} \mu_i$. 

As highlighted by Dipu Kabir et al. (2018), the variation in the outputs of the different networks will be driven by the different random initialization of the weights (parameter uncertainty) and the different bootstrap samples (data uncertainty). Being the bootstrap procedure able to capture both the aleatoric and epistemic uncertainties, it provides more accurate prediction intervals than other methods (i.e., delta method) as also shown in an extensive simulation study in Tibshirani (1996). 

\subsection[Monte Carlo Dropout (Stochastic Forward Passes)]{Monte Carlo Dropout (Stochastic Forward Passes)}

This subsection introduces an alternative to bootstrap methods to construct prediction intervals in a ReLU DNN setting. In this case we introduce randomness into the DNN prediction by applying Monte Carlo dropout. 

Training with dropout (\textit{dropout training} - Figure \ref{fig1}) implies that for each iteration of the learning algorithm different random sub-networks (or \textit{thinned} networks) will be trained.\footnote{Warde-Farley et al. (2014) explain how each sub-network is usually trained for only one gradient step.} Let $h_{zn}$ denote the elements of the vector $\mathbf{h}_{n}$ for a given node $z=1,\ldots,Z_n$ and layer $n=1,\ldots,N$. Srivastava et al. (2014) develop a dropout methodology that is applied to each function $h_{zn}$ to obtain a transformed variable $\overline{h}_{zn}$. This variable is obtained by pre-multiplying $h_{zn}$ by a random variable $r_{zn}$ with distribution function $F(r_{zn}),$ such that $\overline{h}_{zn}=r_{zn}\cdot h_{zn}$, for all $(z,n),$ prior to being fed forward to the activation function of the next layer, $h_{zn+1}$, for all $z=1...Z_{n+1}.$ For any layer $n$, $\mathbf{r}_{n}$ is then a vector of independent random variables, $\mathbf{r}_{n}=[r_{1n},...,r_{Z_{n}n}]\in  \mathbb{R}^{Z_{n}}$. In this paper we consider only the Bernoulli probability distribution $F(r_{zn}),$ where each $r_{zn}$ has probability $p$ of being $1$ (and $q = 1-p$ of being $0$). The vector $\mathbf{r}_{n}$ is then sampled and multiplied element-wise with the outputs of that layer, $h_{zn}$, to create the thinned outputs, $\overline{h}_{zn}$, which are then used as input to the next layer, $h_{zn+1}$. When this process is applied at each layer $n=1...N$, this amounts to sampling a sub-network from a larger network at each forward pass (or gradient step). At test time, the weights are scaled down as $\overline{\mathbf{W}}^{n}$ $=p\mathbf{W}^{n},n=1...N$, returning a deterministic output. We then identify $\mathbf{r}^{\star} =[\mathbf{r}_{1},...,\mathbf{r}_{N}]$ as the collection of independent random variables applied to a feedforward neural network of depth $N+1$\footnote{In practice, an inverted dropout methodology is applied when implementing this methodology in Keras for RStudio. In this case, instead of scaling-down the weights at test time, the weights are scaled-up during train time as $\overline{\mathbf{W}}^{n}$ $=(1/p)\mathbf{W}^{n}, n=1...N$. At test time, a single deterministic forward pass on the unscaled weights $\mathbf{W}^{n}$ is performed.}. Figure \ref{fig1} shows how the dropout mask works; at each training step (forward and backward pass) every neuron of each hidden layer will randomly not be considered when training the network and thus be ''dropped out'' (Géron, 2019).

\begin{figure}[H]
\begin{center}
\begin{tikzpicture}[scale=2]
 
    \node (l1label) at  (0, -2.3) {$\mathbf{x}$};
    \node (l2label) at  (1.5, -2.3) {$\mathbf{h}_{1}$};
    \node (l3label) at  (3,-2.3) {$\mathbf{h}_{2}$};
    \node (l4label) at  (4.5, -2.3) {$\cdots$};
    \node (l5label) at  (6, -2.3) {$\mathbf{h}_{N}$};
    \node (l5label) at  (7, -2.3) {$O$};
    \node (r)[input]   at (0, 1) {$x_{1}$};
    \node (g)[input] at (0, 0) {$x_{2}$};
    \node (b)[input]  at (0,-1) {$x_{3}$};

    \node (h11)[hidden] at (1.5, 1.5) {};
    \node (h12)[hidden] at (1.5, 0.5) {};
    \node[circle,inner sep=0] (h13) at (1.5,-0.5) {\raisebox{5pt}{$\vdots$}};
    \node (h14)[hidden] at (1.5,-1.5) {};
    \node (h21)[hidden] at (3, 1.5) {};
    \node (h22)[hidden, dashed] at (3, 0.5) {};
    \node (h23) at (3,-0.5) {$\vdots$};
    \node (h24)[hidden] at (3,-1.5) {};
    
     \node (h31)at (4.5, 1.5) {$\vdots$};
    \node (h32) at (4.5, 0.5) {$\vdots$};
    \node (h33) at (4.5,-0.5) {$\vdots$};
    \node (h34) at (4.5,-1.5) {$\vdots$};
    
    \node (h41)[hidden,dashed] at (6, 1.5) {};
    \node (h42)[hidden] at (6, 0.5) {};
    \node (h43) at (6,-0.5) {$\vdots$};
    \node (h44)[hidden, dashed] at (6,-1.5) {};

    \node (o1)[output] at (7,0) {};

    \draw[stateTransition] (r) -- (h11) node [midway,above=-0.06cm] {};
    \draw[stateTransition] (r) -- (h12) node [midway,above=-0.06cm] {};
    \draw[stateTransition] (r) -- (h13) node [midway,above=-0.06cm] {};
    \draw[stateTransition] (r) -- (h14) node [midway,above=-0.06cm] {};
    \draw[stateTransition] (g) -- (h11) node [midway,above=-0.06cm] {};
    \draw[stateTransition] (g) -- (h12) node [midway,above=-0.06cm] {};
    \draw[stateTransition] (g) -- (h13) node [midway,above=-0.06cm] {};
    \draw[stateTransition] (g) -- (h14) node [midway,above=-0.06cm] {};
    \draw[stateTransition] (b) -- (h11) node [midway,above=-0.06cm] {};
    \draw[stateTransition] (b) -- (h12) node [midway,above=-0.06cm] {};
    \draw[stateTransition] (b) -- (h13) node [midway,above=-0.06cm] {};
    \draw[stateTransition] (b) -- (h11) node [midway,above=-0.06cm] {};

    \draw[stateTransition] (h11) -- (h21) node [midway,above=-0.06cm] {};
    \draw[stateTransition,dashed] (h11) -- (h22) node [midway,above=-0.06cm] {};
    \draw[stateTransition] (h11) -- (h23) node [midway,above=-0.06cm] {};
    \draw[stateTransition] (h11) -- (h24) node [midway,above=-0.06cm] {};
    \draw[stateTransition] (h12) -- (h21) node [midway,above=-0.06cm] {};
    \draw[stateTransition, dashed] (h12) -- (h22) node [midway,above=-0.06cm] {};
    \draw[stateTransition] (h12) -- (h23) node [midway,above=-0.06cm] {};
    \draw[stateTransition] (h12) -- (h24) node [midway,above=-0.06cm] {};
    \draw[stateTransition] (h13) -- (h21) node [midway,above=-0.06cm] {};
    \draw[stateTransition,dashed] (h13) -- (h22) node [midway,above=-0.06cm] {};
    \draw[stateTransition] (h13) -- (h23) node [midway,above=-0.06cm] {};
    \draw[stateTransition] (h13) -- (h24) node [midway,above=-0.06cm] {};
    \draw[stateTransition] (h14) -- (h21) node [midway,above=-0.06cm] {};
    \draw[stateTransition,dashed] (h14) -- (h22) node [midway,above=-0.06cm] {};
    \draw[stateTransition] (h14) -- (h23) node [midway,above=-0.06cm] {};
    \draw[stateTransition] (h14) -- (h21) node [midway,above=-0.06cm] {};

    \draw[stateTransition] (h21) -- (h31) node [midway,above=-0.06cm] {};
    \draw[stateTransition] (h21) -- (h32) node [midway,above=-0.06cm] {};
    \draw[stateTransition] (h21) -- (h33) node [midway,above=-0.06cm] {};
    \draw[stateTransition] (h21) -- (h34) node [midway,above=-0.06cm] {};
    \draw[stateTransition,dashed] (h22) -- (h31) node [midway,above=-0.06cm] {};
    \draw[stateTransition,dashed] (h22) -- (h32) node [midway,above=-0.06cm] {};
    \draw[stateTransition,dashed] (h22) -- (h33) node [midway,above=-0.06cm] {};
    \draw[stateTransition,dashed] (h22) -- (h34) node [midway,above=-0.06cm] {};
    \draw[stateTransition] (h23) -- (h31) node [midway,above=-0.06cm] {};
    \draw[stateTransition] (h23) -- (h32) node [midway,above=-0.06cm] {};
    \draw[stateTransition] (h23) -- (h33) node [midway,above=-0.06cm] {};
    \draw[stateTransition] (h23) -- (h34) node [midway,above=-0.06cm] {};
    \draw[stateTransition] (h24) -- (h31) node [midway,above=-0.06cm] {};
    \draw[stateTransition] (h24) -- (h32) node [midway,above=-0.06cm] {};
    \draw[stateTransition] (h24) -- (h33) node [midway,above=-0.06cm] {};
    \draw[stateTransition] (h24) -- (h31) node [midway,above=-0.06cm] {};

    \draw[stateTransition] (h31) -- (h41) node [midway,above=-0.06cm] {};
    \draw[stateTransition] (h31) -- (h42) node [midway,above=-0.06cm] {};
    \draw[stateTransition] (h31) -- (h43) node [midway,above=-0.06cm] {};
    \draw[stateTransition,dashed] (h31) -- (h44) node [midway,above=-0.06cm] {};
    \draw[stateTransition] (h32) -- (h41) node [midway,above=-0.06cm] {};
    \draw[stateTransition] (h32) -- (h42) node [midway,above=-0.06cm] {};
    \draw[stateTransition] (h32) -- (h43) node [midway,above=-0.06cm] {};
    \draw[stateTransition,dashed] (h32) -- (h44) node [midway,above=-0.06cm] {};
    \draw[stateTransition] (h33) -- (h41) node [midway,above=-0.06cm] {};
    \draw[stateTransition] (h33) -- (h42) node [midway,above=-0.06cm] {};
    \draw[stateTransition] (h33) -- (h43) node [midway,above=-0.06cm] {};
    \draw[stateTransition,dashed] (h33) -- (h44) node [midway,above=-0.06cm] {};
    \draw[stateTransition] (h34) -- (h41) node [midway,above=-0.06cm] {};
    \draw[stateTransition] (h34) -- (h42) node [midway,above=-0.06cm] {};
    \draw[stateTransition] (h34) -- (h43) node [midway,above=-0.06cm] {};
    \draw[stateTransition] (h34) -- (h41) node [midway,above=-0.06cm] {};
    
    \draw[stateTransition,dashed] (h41) -- (o1) node [midway,above=-0.06cm] {};
    \draw[stateTransition] (h42) -- (o1) node [midway,above=-0.10cm] {};
    \draw[stateTransition] (h43) -- (o1) node [midway,above=-0.06cm] {};
    \draw[stateTransition,dashed] (h44) -- (o1) node [midway,above=-0.06cm] {};

\end{tikzpicture}
\end{center}
\caption[ReLu Deep Neural Network]{ReLu Deep Neural Network with bias terms $0$ and dropout mask.}
\label{fig1}
\end{figure}
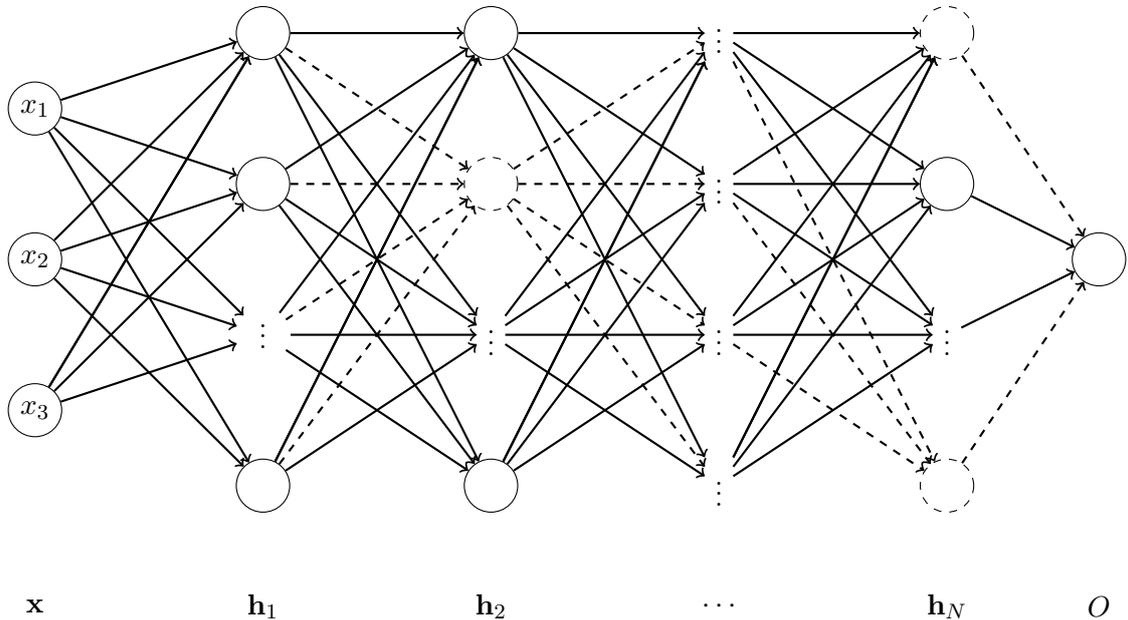

A natural interpretation of this methodology follows from the seminal contribution of Gal and Ghahramani (2016a). These authors develop a new theoretical framework casting dropout training in DNNs as approximate Bayesian inference in deep Gaussian processes. As a byproduct of this theory, Gal and Ghahramani (2016a) provide the tools to model prediction uncertainty with dropout in DNNs. A growing branch of the literature has been focusing on the Bayesian interpretation of dropout\footnote{Hinton et al. (2012) in their seminal paper associate dropout training to a form of Bayesian learning.} (see among others, Gal and Ghahramani (2016a, 2016b) and Kingma et al. (2015)). Maeda (2014) explains how dropout training can be considered an approximate learning method of the model parameters that optimizes a weighted sum of the likelihoods of all possible models. 

Starting from this interpretation, one could consider dropout as a tool for the estimation of the posterior of a Bayesian neural network. More specifically, let $p(\widehat{y} \ | \mathbf{x}, \mathbf{X},\mathbf{Y})$ denote the distribution of the predictive output $\widehat{y}$ conditional on the set of observations $\mathbf{X}=\{\mathbf{x}_1,\ldots,\mathbf{x}_n\}$ and $\mathbf{Y}=\{y_1,\ldots,y_n\}$. The predictive probability distribution of the DNN model is 
\begin{equation} \label{predBay}
p(\widehat{y} \ | \mathbf{x}, \mathbf{X},\mathbf{Y}) = \int_{\bm{\Omega}} p(\widehat{y} \ | \mathbf{x}, \bm{\omega}) p(\bm{\omega} \ | \ \mathbf{X},\mathbf{Y}) \text{d}\bm{\omega},
\end{equation} 
with $p(\widehat{y} \ | \mathbf{x}, \bm{\omega})$ the likelihood function of the observations, and $\bm{\omega} \in \bm{\Omega}$ where $\bm{\Omega}$ denotes the parameter space. The posterior probability distribution $p(\bm{\omega} \ | \ \mathbf{X},\mathbf{Y})$ is intractable. 

Gal and Ghahramani (2016a) propose DNN dropout to approximate this distribution. More formally, under model dropout, we consider a distribution function $q(\bm{\omega})$ that follows a Bernoulli distribution, $\text{Ber}(p)$. The above predictive distribution in this Bayesian neural network setting can be approximated by 
\begin{equation} \label{predBay}
p(\widehat{y} \ | \mathbf{x}, \mathbf{X},\mathbf{Y}) = \int_{\Omega} p(\widehat{y} \ | \mathbf{x}, \bm{\omega}) q(\bm{\omega})\text{d}\bm{\omega}.
\end{equation} 

In practice this predictive distribution can be approximated using Monte Carlo methods. Thus, by sampling $T$ sets of vectors from the Bernoulli distribution $\{\mathbf{r}^{\star(t)}\}_{t=1}^{T}$, one can approximate the above predictive distribution from the random sample $\widehat{y}(\mathbf{x}_i;\bm{\widehat{\omega}}^{(t)})$, for $i=1,\ldots,n$, where $\bm{\widehat{\omega}}^{(t)} = \{\widehat{\mathbf{W}}^{1(t)},\ldots,\widehat{\mathbf{W}}^{N(t)},\widehat{b}_1^{(t)},\ldots,\widehat{b}_N^{(t)}\}$ denotes the sequence of weights associated to the different nodes and layers of the neural network and the associated bias parameters for a given pass $t$ for $t=1,\ldots,T$. 

Using this Monte Carlo dropout technique, Gal and Ghahramani (2016a) propose the first moment from the MC predicted outputs as the model prediction:

\begin{equation}\label{gal1}
\bar{f}_{MC}(\mathbf{x}_i) = \frac{1}{T}\sum_{t=1}^{T} \widehat{y}(\mathbf{x}_i;\bm{\widehat{\omega}}^{(t)}), \ \text{for} \ i=1,\ldots,n.
\end{equation}
These authors show that, in practice, this is equivalent to performing T stochastic forward passes through the network and averaging the results. This result has been presented in the literature before as model averaging. Srivastava et al. (2014) have reasoned empirically that MC dropout can be approximated by averaging the weights of the network (multiplying each weight $\mathbf{W}^{n}$ by $p$ at test time, and referred to as standard dropout). 

Importantly, the model parameters $\bm{\omega}$ are fixed across random samples implying that the cross-correlation between the predictions $\widehat{y}(\mathbf{x}_i;\bm{\widehat{\omega}}^{(t)})$ and $\widehat{y}(\mathbf{x}_i;\bm{\widehat{\omega}}^{(t')})$ for $t,t'=1,\ldots,T$ is perfect. Then, the predictive variance is defined as
\begin{equation} \label{varBay0}
\sigma_{MC}^2 = \sigma_{e}^2 + \frac{1}{T^2} \sum_{t=1}^{T} \sum_{t'=1}^{T} E \left[\left(\widehat{y}(\mathbf{x}_i;\widehat{\bm{\omega}}^{(t)}) - E[\widehat{y}(\mathbf{x}_i;\widehat{\bm{\omega}}^{(t)})] \right) \left(\widehat{y}(\mathbf{x}_i;\widehat{\bm{\omega}}^{(t')}) - E[\widehat{y}(\mathbf{x}_i;\widehat{\bm{\omega}}^{(t')})] \right) \right],
\end{equation}
The first component on the right hand side expression of \eqref{varBay0} captures the aleatoric uncertainty whereas the second term captures the epistemic uncertainty associated to parameter estimation. The second term includes the estimation of the variance and covariance terms between the different random samples obtained from using dropout. Thus, under the assumption that the approximation error is negligible, the above predictive variance can be estimated as
\begin{equation} \label{varBay00}
\widehat{\sigma}_{MC}^2 = \widehat{\sigma}_{e}^2 + \frac{1}{T} \sum_{t=1}^{T}  \left(\widehat{y}(\mathbf{x}_i;\widehat{\bm{\omega}}^{(t)}) - \bar{f}_{MC}(\mathbf{x}_i) \right)^2,
\end{equation}
with $\widehat{\sigma}_{e}^2 = \frac{1}{n}\sum_{i=1}^{n} \left(y_{i} - \bar{f}_{MC}(\mathbf{x}_{i}) \right)^2$ a consistent estimator of $\sigma_{e}^2$ under homoscedasticity of the error term, see also Gal and Ghahramani (2016a) and Kendall and Gal (2017). A suitable prediction interval for $y_i$ under the assumption that $p(\widehat{y} \ | \mathbf{x}, \bm{\omega})$ is normally distributed is  
\begin{equation} \label{pred4}
\bar{f}_{MC}(\mathbf{x}_i) \pm z_{1-\alpha/2} \widehat{\sigma}_{MC}.
\end{equation}

Recent literature focusing on the approximation of the predictive distribution of DNNs has proposed several algorithms - based on the MC dropout of Gal and Ghahramani (2016a) - for the estimation of the prediction uncertainty in deep learning. As an example, Serpell et al. (2019) augment the MC dropout by implementing the MVE discussed above and stochastic forward passes. If the MVE approach allows modeling the data uncertainty - accommodating a varying $e$, the Monte Carlo dropout captures the uncertainty in the model parameters. The two procedures together allow the correct estimation of $\sigma_{MC}^2$; Zhu and Laptev (2017) improve over the original Monte Carlo dropout by estimating the noise level using the residual sum of squares evaluated on a hold-out set\footnote{The authors precise that the approach of Gal and Ghahramani (2016a) relies on the implausible assumption of knowing the correct noise level a priori.} - Equation \ref{varBay00}.

Finally, the present paper highlights three important aspects related to the MC dropout originally proposed by Gal and Ghahramani (2016a). First, it is possible to extend the original approach to the approximation of the predictive distribution of deep neural networks outside a Bayesian framework (see Cortes-Ciriano and Bender, 2019). As one could notice, using dropout also at test phase allows randomizing the output of the DNN at each forward pass and thus, by performing $T$ stochastic forward passes, it is possible to obtain the sample $\{\widehat{y}(\mathbf{x}_{i}; \widehat{\bm{\omega}}^{(t)})\}_{t=1}^{T}$. Second, if the MC dropout is implemented outside a Bayesian framework, it is pivotal to tune the dropout rate on a test sample and not on a train sample; in fact, as suggested by Lakshminarayanan et al. (2017), tuning the dropout rate on the training data implies interpreting dropout as a tool for Bayesian inference (any Bayesian posterior should be approximated starting only from the training data). Last but not least, the estimation of the $\sigma_{MC}^2$ depends on the choice of $p$. As the epistemic uncertainty in the MC dropout is determined solely by the choice of $p$, if $p$ is set equal to $1$ the epistemic uncertainty measured by $\frac{1}{T} \sum_{t=1}^{T} \left(\widehat{y}(\mathbf{x}_i;\widehat{\bm{\omega}}^{(t)}) - \bar{f}_{MC}(\mathbf{x}_i) \right)^{2}$ will be zero. Thus, the empirical coverage rate of the MC dropout will depend significantly upon the right choice of $p$ (as opposed to the other analyzed methods). This final insight is also confirmed by the simulation results of Levasseur et al. (2017).

\section[Extra-neural networks (Fixed Bernoulli Mask)]{Extra-neural networks (Fixed Bernoulli Mask)} \label{predictionnovel}

In this section we introduce a novel ensemble predictor within deep neural network models. This methodology also allows us to construct prediction intervals based upon the work of Srivastava et al. (2014). In this case the original concept of ensemble of sub-networks - from which the dropout training is built upon - is adopted. The Bernoulli mask $\mathbf{r}^{\star}$ introduces an additional randomization scheme to the predictions obtained from the ensemble of neural networks that ensures independence of the individual predictor models. 

For notation purposes, we will identify the fixed Bernoulli mask as $\mathbf{\bar{r}}^{\star}$ as opposed to $\mathbf{r}^{\star}$ used in dropout training. In other words, $T$ sets of vectors $\{\mathbf{\bar{r}}^{\star(t)}\}_{t=1}^{T}$ are sampled from the Bernoulli distribution prior to training (instead of test time with Monte Carlo dropout) that are kept constant during both train and test phases. This approach reduces to train and independently fit $T$ random sub-networks on the same dataset. In this setting, generating the predictive distribution is similar, in spirit, to an ensemble approach that trains different sub-neural networks on the same dataset. The proposed algorithm - being based on the extremely randomized trees proposed by Geurts et al. (2006) - is called \textit{extra-neural networks}.

Let $\bar{f}_{EN}(\mathbf{x}_i)$ denote the ensemble predictor obtained from the extra-neural networks approach that is constructed as
\begin{equation}\label{ensembleoutput100}
\bar{f}_{EN}(\mathbf{x}_i) = \frac{1}{T}\sum_{t=1}^{T} f_{t}(\mathbf{x}_{i}; \widehat{\bm{\omega}}^{(t)}), \ \text{for} \ i=1,\ldots,n.
\end{equation}
We consider $T$ fitted sub-networks defined as $f_{t}(\mathbf{x}_{i}; \widehat{\bm{\omega}}^{(t)})$ with $t = 1, \cdots, T$. We use $f_{t}$ to note that each prediction belongs to a potentially different neural network model; $\widehat{\bm{\omega}}^{(t)}$ denotes the parameter estimates obtained from fitting each sub-network independently. 

\vspace{2mm}

Before analyzing the prediction intervals for the extra-neural network, it is convenient to analyze the factors that influence the prediction accuracy of the model. To do this, we compute the mean square prediction error (MSPE) of the prediction conditional on the regressor vector $\mathbf{x}_i$. 
Then,
\begin{equation}\label{mseEnsemble}
MSPE(\bar{f}_{EN}(\mathbf{x}_i)) \equiv \mathbb{E}[(\bar{f}_{EN}(\mathbf{x}_i) - y_{i})^{2}] = \text{Bias}^{2}(\bar{f}_{EN}(\mathbf{x}_i)) + V(\bar{f}_{EN}(\mathbf{x}_i)).
\end{equation}
We compute the conditional bias and variance of $\bar{f}_{EN}(\mathbf{x}_i)$ as 

\begin{equation}
\text{Bias}(\bar{f}_{EN}(\mathbf{x}_i)) \equiv \mathbb{E}[\bar{f}_{EN}(\mathbf{x}_i) - y_i] = \frac{1}{T}\sum_{t=1}^{T}\mathbb{E}[f_{t}(\mathbf{x}_{i}; \widehat{\bm{\omega}}^{(t)})] - f(\mathbf{x}_{i}), 
\end{equation}
and
\begin{equation*}
V(\bar{f}_{EN}(\mathbf{x}_i)) \equiv \mathbb{E} \left(\bar{f}_{EN}(\mathbf{x}_i) - \mathbb{E}[\bar{f}_{EN}(\mathbf{x}_i)] \right)^2 = \mathbb{E} \left[\bar{f}_{EN}^2(\mathbf{x}_i)\right] - \mathbb{E}^2[\bar{f}_{EN}(\mathbf{x}_i)],
\end{equation*}

\noindent such that
\begin{equation*}
V(\bar{f}_{EN}(\mathbf{x}_i)) = \frac{1}{T^2}\sum_{t=1}^{T} \sum_{t'=1}^{T} \left(\mathbb{E}[f_{t}(\mathbf{x}_{i}; \widehat{\bm{\omega}}^{(t)}) f_{t'}(\mathbf{x}_{i}; \widehat{\bm{\omega}}^{(t')})] -  \mathbb{E}[f_{t}(\mathbf{x}_{i}; \widehat{\bm{\omega}}^{(t)})] \mathbb{E}[f_{t'}(\mathbf{x}_{i}; \widehat{\bm{\omega}}^{(t')})] \right).
\end{equation*}
Furthermore, assuming that the first two statistical moments of all the individual predictors indexed by $t=1,\ldots,T$ are equal, with $\mathbb{E} \left[f_{t}(\mathbf{x}_{i}; \widehat{\bm{\omega}}^{(t)})\right] = f(\mathbf{x}_i) + \mu_i$, where $\mu_i$ is the bias term, $\mathbb{V} \left[f_{t}(\mathbf{x}_{i}; \widehat{\bm{\omega}}^{(t)}) \right] = \sigma^{2}_{\widehat{\bm{\omega}}}(\mathbf{x}_i)$, and Cov$ \left[f_{t}(\mathbf{x}_{i}; \widehat{\bm{\omega}}^{(t)}) f_{t'}(\mathbf{x}_{i}; \widehat{\bm{\omega}}^{(t')})\right] = c_i$, we obtain

\begin{equation}\label{refMSE}
MSPE(\bar{f}_{EN}(\mathbf{x}_i)) = \mu_i^{2} + \frac{1}{T} \sigma^{2}_{\widehat{\bm{\omega}}}(\mathbf{x}_i) + \frac{T-1}{T} c_i.
\end{equation}
This expression extends Zhou (2012) by showing that the MSPE of the ensembler \eqref{ensembleoutput100} depends on the variance of the individual ensemblers, their covariance and the approximation bias. The smaller the covariance, the smaller the generalization error of the ensemble. In contrast, if the different predictors are perfectly correlated (as for the MC dropout) we know that $c_i=\sigma^{2}_{\widehat{\bm{\omega}}}(\mathbf{x}_i)$ and thus $MSPE(\bar{f}_{EN}(\mathbf{x}_i)) = \sigma^{2}_{\widehat{\bm{\omega}}}(\mathbf{x}_i)$ - effectively reducing to zero the effect of ensembling. Similarly, the MSPE is minimized when the errors are perfectly uncorrelated and thus when $c_i = 0$. 

This result has important implications when analyzing the epistemic uncertainty of an extra-neural network. If it is assumed that the correlation among the predictions from the sub-networks is equal to zero, then as $T \to \infty$, the $MSPE(\bar{f}_{EN}(\mathbf{x}_i))$ converges to zero, assuming that the approximation bias is negligible. Therefore, a suitable prediction interval is

\begin{equation}\label{predictionEnsemble1} 
\bar{f}_{EN}(\mathbf{x}_i) \pm z_{1-\alpha/2} \left( \frac{\widehat{\sigma}^{2}_{\widehat{\bm{\omega}}}(\mathbf{x}_i)}{T} + \widehat{\sigma}_{e}^{2} \right)^{1/2},
\end{equation}
\vspace{2mm}
\noindent
with $\widehat{\sigma}^{2}_{\widehat{\bm{\omega}}}(\mathbf{x}_i) = \frac{1}{T} \sum_{t=1}^{T} (f_{t}(\mathbf{x}_{i};\bm{\widehat{\omega}}^{(t)}) - \bar{f}_{EN}(\mathbf{x}_i))^2$ and $\widehat{\sigma}_{e}^{2} = \frac{1}{n} \sum_{i=1}^{n} \left(y_{i} - \bar{f}_{EN}(\mathbf{x}_i) \right)^2$, where $n$ is the length of the test sample.\footnote{Note that for obtaining a consistent estimator of $\widehat{\sigma}_{e}^{2}$ we have imposed homoscedasticity of the error terms $\epsilon_i$ over the test sample.} 

As explained in Zhou (2012), the covariance term in equation \eqref{refMSE} captures the diversity existing among the $T$ different sub-networks identifying the extra-neural network. The aim of the extra-neural network approach proposed in this paper is to construct individual predictors that are mutually independent such that the prediction interval \eqref{predictionEnsemble1}  is valid. The diversity in the model predictions depends on the variance of the Bernoulli masks generated by the random sample $\{\mathbf{\bar{r}}^{\star(t)}\}_{t=1}^{T}$. It is well known that the variance of a Bernoulli distribution is defined as $\varsigma^{2} = p(1-p)$; therefore, it can be easily shown that the solution to $\partial \varsigma^{2} /\partial p = 0$ is $p = 1/2$ and that $\partial^{2} \varsigma^{2} / \partial p^{2} = -2$ showing that $\varsigma^{2}$ is maximized in $p = q = 0.5$. Consequentially, one could conclude that the covariance in Equation \ref{mseEnsemble} is minimized for $p = 0.5$ and maximized for $p = 0$ and $p = 1$, see also Figure \ref{bernoullivariance}:

\begin{figure}[H]
\captionsetup{singlelinecheck = false, justification=justified} 
\centering
{\includegraphics[page = 1, clip, trim=0cm 19cm 0cm 0cm, width=1\textwidth]{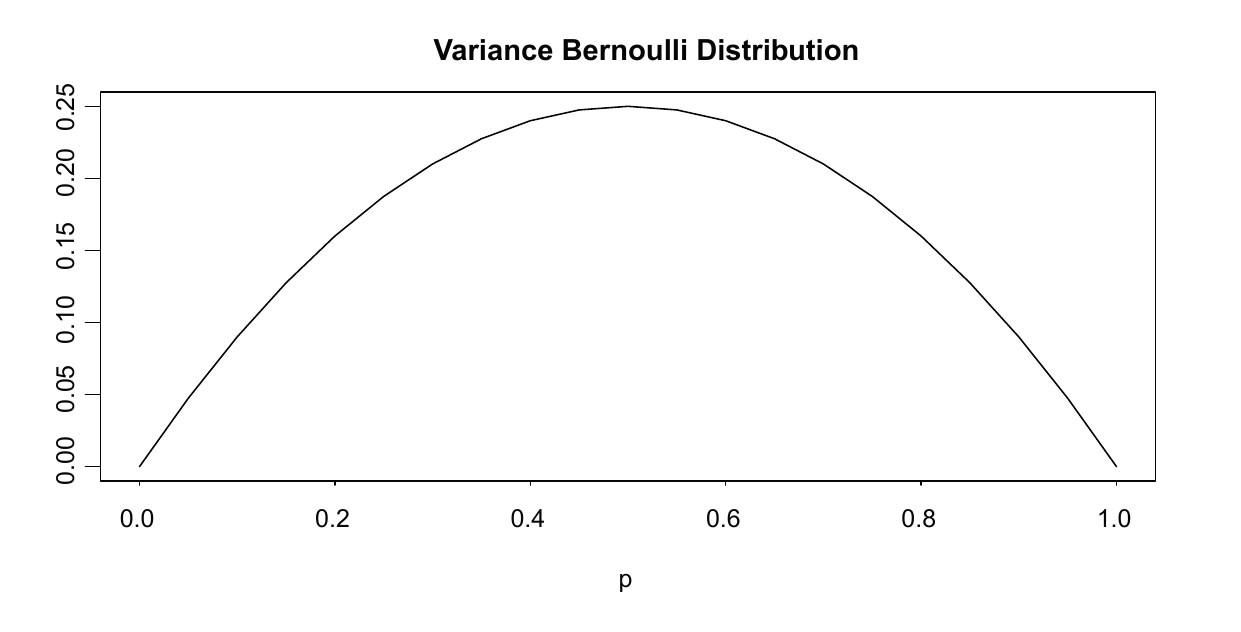}}
\caption[Bernoulli Variance]{Bernoulli Variance. The variance ($\varsigma^{2}$) is maximized for $p = 0.5$ and minimized ($=0$) for $p = 0$ and $1$.}
\label{bernoullivariance}
\end{figure}

However, a complete analysis of the covariance of an ensemble of neural networks must consider the relation existing between the number of hidden nodes and the particular data generating process analyzed. Based on the literature on approximation theory and DNNs, the number of hidden nodes defines the approximation power (or flexibility) of the neural networks (for a summary on the topic see Calvo-Pardo et al., 2020). Farrell et al. (2019) - by comparing DNN structures to different nonparametric techniques for approximating unknown continuous functions - also make explicit the dependence between the number of hidden nodes in the DNN ($Z_{\text{tot}}$) and the approximation power. Therefore, if the size of the networks is such that the \textit{ambiguity} - measure of disagreement among the different networks on a specific input, see Krogh and Vedelsby (1995) for a detailed analysis - is too low, the assumption of $c = 0$ becomes unrealistic. 

Based on the above paragraph, one could conclude that the analysis of the covariance in an extra-neural network must consider not only $p$ that determines the variance of $\{\mathbf{\bar{r}}^{\star(t)}\}_{t=1}^{T}$ but also the particular data generating process under study, as $Z_{\text{dropout}}$ is also determined by $p$. As the two effects must be considered together when choosing the probability $p$, one must consider that $p$ converging to $0.5$ from above or from below may have a similar impact in terms of decrease in $c$ but and opposite effect on the dimension of $Z_{\text{dropout}}$. As $p$ converges to $0.5$ from below, the dimensions of the sub-networks will increase (higher probability for each neuron to be $1$ and thus to be retained in the sub-network). Conversely, $p$ converging to $0.5$ from above will ensure a reduction of the number of hidden nodes in the $T$ sub-networks (higher probability of being ''dropped out'' - $q = 1-p$). 

\begin{algorithm}[H]
\caption{Extra-neural networks}\label{alg1}
\begin{flushleft}
       \textbf{INPUT:} Training Data $\{x_i^{\llcorner} \equiv (\mathbf{x}_i,y_i) \}_{i=1}^{M}$  \\
       \textbf{OUTPUT:} Prediction Interval $\widehat{f}(\mathbf{x};\bm{\omega})$.
\end{flushleft}

\begin{algorithmic}[1]
\Procedure{T learners}{}\\

\State Define depth and width of \textit{original} neural network. 

\vspace{4mm}
\While{(t < T)}
\State Generate a Bernoulli mask $\mathbf{\bar{r}}^{\star}$ prior to training.
\State Apply Bernoulli mask $\mathbf{\bar{r}}^{\star}$ to the \textit{original} neural network.
\State Train random thinned network on $\mathbf{x}^{\llcorner}$ with random initialization of $\{\mathbf{W}_{0}^{n}\}_{n=1}^{N}$
\State Trained thinned network $\rightarrow$ Deterministic forward pass on test data. 
\State Store $f_{t}(\mathbf{x}_{i}; \widehat{\bm{\omega}}^{(t)})$.
\EndWhile

\vspace{3mm}
\State Compute the ensemble estimate: 

\begin{equation}\label{extranew1}
\bar{f}_{EN}(\mathbf{x}_i) = \frac{1}{T}\sum_{t=1}^{T} f_{t}(\mathbf{x}_{i}; \widehat{\bm{\omega}}^{(t)})
\end{equation}

\State Compute the epistemic and aleatoric variance: 
\begin{equation}\label{extranew2}
\begin{cases}
\widehat{\sigma}^{2}_{\widehat{\bm{\omega}}}(\mathbf{x}_i) = \frac{1}{T} \sum_{t=1}^{T} [f_{t}(\mathbf{x}_{i};\bm{\widehat{\omega}}^{(t)}) - \bar{f}_{EN}(\mathbf{x}_i)]^2 \\
\widehat{\sigma}_{e}^{2} = \frac{1}{n} \sum_{i=1}^{n} \left(y_{i} - \bar{f}_{EN}(\mathbf{x}_i) \right)^2
\end{cases}
\end{equation}

\State Define Prediction Interval: 
\begin{equation}\label{PIEN}
\bar{f}_{EN}(\mathbf{x}_i)  \pm z_{1-\alpha/2} \widehat{\sigma}_{\epsilon},
\end{equation}
with $\widehat{\sigma}_{\epsilon} = \left( \frac{\widehat{\sigma}^{2}_{\widehat{\bm{\omega}}}(\mathbf{x}_i)}{T} + \widehat{\sigma}_{e}^{2} \right)^{1/2}$.

\vspace{4mm}
\Return{Prediction interval (\ref{PIEN})}
\EndProcedure
\end{algorithmic}
\end{algorithm}

\bigskip

Algorithm \ref{alg1} reports the procedure to be used for implementing the Extra-neural network. In order to generate $\{f_{t}(\mathbf{x};\widehat{\bm{\omega}}^{(t)})\}_{t=1}^{T}$, we sample $T$ vectors $\{\mathbf{\bar{r}}^{\star(t)}\}_{t=1}^{T}$ prior to training. Each fixed Bernoulli mask is applied independently to the original network returning $T$ independent sub-networks of size $Z^{(t)}_{\text{dropout}} \leq Z_{\text{tot}}$. Each sub-network is then trained independently on $\mathbf{x}^{\llcorner}$, and $T$ deterministic forward passes are performed at test phase. Thus, even if the novel algorithm is based upon the original idea of dropout proposed by Srivastava et al. (2014) and introduces randomness by means of a random sample $\{\mathbf{\bar{r}}^{\star(t)}\}_{t=1}^{T}$, it is closer to classical ensemble methods than to training with dropout. This has important implications while training. In this case, performing weight scaling at test phase (or train phase if the algorithm is implemented in Keras) is not required as the Bernoulli mask is applied before training. Training $T$ independent sub-networks identified by $\{\mathbf{\bar{r}}^{\star(t)}\}_{t=1}^{T}$ makes no longer necessary to ensure that the expected total input to the units of a DNN at test time is approximately the same as the expected value at training (see Goodfellow et al., 2016). 

The procedure reported in Algorithm \ref{alg1} shows that an extra-neural network is an ensemble of $T$ neural networks with randomized weights and structures and no data resampling. Based on the results reported by Pearce et al. (2018), Lee et al. (2015) and Lakshminarayanan et al. (2017) regarding deep ensembles\footnote{Deep ensembles and ensembles of DNNs are considered synonym for the rest of the paper.} it is expected that the extra-neural network algorithm will improve over a bootstrapping ensemble approach. More precisely, Lee et al. (2015) show how parameter resampling without bootstrap resampling - equivalent to training T different $f(\mathbf{x}_{i}; \widehat{\bm{\omega}}^{(t)})$ on $\mathbf{x}^{\llcorner}$ - outperforms a bootstrap approach (analyzed in Subsection \ref{bootsubsection}) in terms of predictive accuracy; Lakshminarayanan et al. (2017) complement the results of Lee et al. (2015) by showing that data resampling in deep ensembles deteriorates not only the prediction accuracy but also the definition of the predictive uncertainty of the ensemble itself. 

Therefore, the extra-neural networks by randomizing not only the weights of the $T$ sub-networks but also their structure, and by fitting the networks on the entire training set $\{\mathbf{x}_i\}_{i=1}^{M}$, are expected to outperform the bootstrap approach in terms of both out-of-sample prediction accuracy (Lee et al., 2015) and uncertainty quantification (Lakshminarayanan et al., 2017)\footnote{By considering deep ensembles the equivalent of a random forest (Breiman, 2001) where the single learners are neural networks and where the parameter uncertainty is captured not by the random subset selection of features at each node (trees) but by random weight initialization, the extra randomization introduced by extra-neural networks is comparable to the extremely randomized trees in Geurts et al. (2006). In this case, randomizing also the structure is equivalent to randomizing the cut-point at each node in a tree.}. 

The main drawback of the extra-neural network algorithm is associated to the computing power required. In particular, if the computational requirements of the proposed methodology are equivalent to existing bootstrapping procedures (with and without data resampling), they are significantly greater than the ones of the MC dropout methodology. However, due to the parameter sharing in the MC dropout, the extra-neural networks will ensure a lower MSPE (see equation \eqref{refMSE}). Additionally, it is expected an improvement also in terms of hyperspace: the novel methodology allows reaching a good estimation performance without the pivotal fine-tuning that is required by the other procedures. As in the case of bootstrap based procedures, the independence among the different learners in the extra-neural networks allows parallel computing ensuring savings in computational time. Last but not least, the extra-neural network improves over a bootstrap based approach in terms of applicability: if the bootstrap approach relies on the assumption of $i.i.d$ observations, the extra-neural network does not. 

All the results analyzed in Section \ref{prediction} and \ref{predictionnovel} will be formally evaluated in an extensive simulation study focused on assessing if the reported procedures return correct prediction intervals (empirical coverage close to the nominal one) for different significance levels and data generating processes. Finally, the empirical experimental setting of Hern\'{a}ndez-Lobato and Adams (2015) is implemented to compare the performance (in terms of RMSPE) of the different algorithms. 

\section[Monte Carlo simulation]{Monte Carlo simulation}\label{montecarlo}
The aim of this simulation section is twofold. First, we assess the accuracy of the pointwise predictions of the above ensemble prediction models, and second, we study the empirical coverage probability of the interval forecasts associated to each prediction model.

We analyze the empirical coverage rates of the prediction intervals obtained from expression \eqref{pred2d} (bootstrap approach), expression \eqref{pred4} (MC dropout) and expression \eqref{predictionEnsemble1} (extra-neural network)\footnote{The authors acknowledge the use of the IRIDIS High Performance Computing Facility, and associated support services at the University of Southampton, in the completion of this work.}. For each prediction interval, the empirical coverage rates ($\bar{\alpha}$) for three different significance levels ($0.01$, $0.05$, and $0.10$) are computed. This allows evaluating the correctness of the constructed prediction intervals for different significance levels. All three procedures are analyzed for increasing $T = [30, 50, 70]$, and for a sample size $M+n = 1200+300$. When the small-dimensional linear process is considered -  in order to evaluate the impact that different $p$s may have on the correct definition of the prediction intervals - we will consider $p = [0.995, 0.990, 0.950, 0.900, 0.800]$\footnote{The choice of the dropout rate $q = 1 - p$ is dictated by a really small network size and also a fairly small simulated dataset.}. Subsection \ref{controlSett} reports the setting for the simulation of the small dimensional linear and nonlinear data generating processes; Subsection \ref{normalSett} summarizes the results. Subsection \ref{sim_correlation} presents a simulation study that confirms empirically the absence of correlation between the different individual predictors in the extra-neural network approach.

\subsection[Data Generating Processes]{Data Generating Processes}\label{controlSett}
When the nonlinear data generating process (DGP) is considered, the dataset $\mathbf{x} \in \mathbb{R}^{5}$ is defined by $\mathbf{x}_{1} \sim\mathcal{N}(-4, 1)$, $\mathbf{x}_{2} \sim\mathcal{N}(2, 1)$, $\mathbf{x}_{3} \sim\mathcal{N}(2, 1)$, $\mathbf{x}_{4} \sim\mathcal{N}(2, 1)$, and $\mathbf{x}_{5} \sim\mathcal{N}(4, 1)$\footnote{The means are randomly sampled with replacement from a domain defined in $[-5, 5]$.}. In order to introduce correlation among the variables, the Choleski decomposition is applied. The desired correlation matrix is defined as: 

\vspace{3mm}
\begin{equation}\label{corChole}
\mathbf{C} = 
\begin{bmatrix}
1 & 0.5 & 0.6 & 0.7 & 0.5 \\
0.5 & 1 & 0.7 & 0.8 & 0.5 \\
0.6 & 0.7 & 1 & 0.7 & 0.5 \\
0.7 & 0.8 & 0.7 & 1 & 0.8 \\ 
0.9 & 0.5 & 0.6 & 0.8 & 1 
\end{bmatrix}
\end{equation} 

\vspace{2mm} 
Before imposing the correlation structure in $\mathbf{C}$, it is necessary to make sure that the simulated variables are independent. To do so, the current correlation matrix $\mathbf{\Sigma}$ is calculated; following, the inverse of the Cholesky factorization ($\mathbf{A}^{-1}$) of $\mathbf{\Sigma}$ is computed. By matrix multiplying $\mathbf{A}^{-1}$ and $\mathbf{x}$, we will ensure that the obtained dataset will be defined by independent Normally distributed variables. Finally, the Cholesky factorization ($\mathbf{A}$) of $\mathbf{C}$ is calculated and multiplied by the simulated dataset, ensuring that $\mathbf{Z} = \mathbf{x} \mathbf{A} \approx \mathcal{N}(0, \mathbf{C})$. 

The nonlinear DGP is defined by a ReLu DNN with two hidden layers of width $3$ and $2$ respectively, and bias equal to $1$ across all hidden layers\footnote{A similar DGP is also simulated in Tibshirani (1996) with $\mathbf{x} \in \mathbb{R}^{4}$, and a shallow network with sigmoid activation functions and two hidden nodes; the Gaussian error $\bm{\epsilon}$ follows the same distribution.}: 

\begin{align*}
\mathbf{T}_{1} &= \underbrace{\theta(1 - 3\mathbf{x}_{1} - 2\mathbf{x}_{2} + 1\mathbf{x}_{3} + 5\mathbf{x}_{4} - 3\mathbf{x}_{5})}_{\mathbf{h}_{11}} + \underbrace{\theta(1 + 4\mathbf{x}_{1} + 5\mathbf{x}_{2} + 2\mathbf{x}_{3} + 2\mathbf{x}_{4} - 5\mathbf{x}_{5})}_{\mathbf{h}_{21}} \\ 
& + \underbrace{\theta(1 - 3\mathbf{x}_{1} - 4\mathbf{x}_{2} + 2\mathbf{x}_{3} - 2\mathbf{x}_{4} + 3\mathbf{x}_{5})}_{\mathbf{h}_{31}}\\ 
& \\ 
\mathbf{T}_{2} &=  \underbrace{\theta(1 - 1\mathbf{h}_{11} + 3\mathbf{h}_{21} + 5\mathbf{h}_{31})}_{\mathbf{h}_{12}} +  \underbrace{\theta(1 - 2\mathbf{h}_{11} + 3\mathbf{h}_{21} + 5\mathbf{h}_{31})}_{\mathbf{h}_{22}} \\ 
& \\ 
\mathbf{y} &= 1 + \mathbf{h}_{12} + 2\mathbf{h}_{22} + \bm{\epsilon}\label{cicci}
\end{align*}

\vspace{3mm} 
\noindent
with $\epsilon \sim \mathcal{N}(0, 0.7)$, $\theta(\mathbf{x}) = \max\{0, \mathbf{x}\}$, and the coefficients (network weights) randomly sampled with replacement from $[-5, 5]$. The standard deviation of the error term is set equal to $0.7$ in order to reduce the nuisance in the system by differentiating the stochastic behavior of the regressors $\mathbf{x}$ and of the error term. Figure \ref{visualDGP} provides a visual representation of the underlying DGP and the obtained dependent variable $\mathbf{y}$:

\begin{figure}[H]
\captionsetup{singlelinecheck = false, justification=justified} \centering
\framebox[\textwidth]{\includegraphics[page = 1, clip, trim= 0cm 9cm 0cm 1cm, width= 1\textwidth]{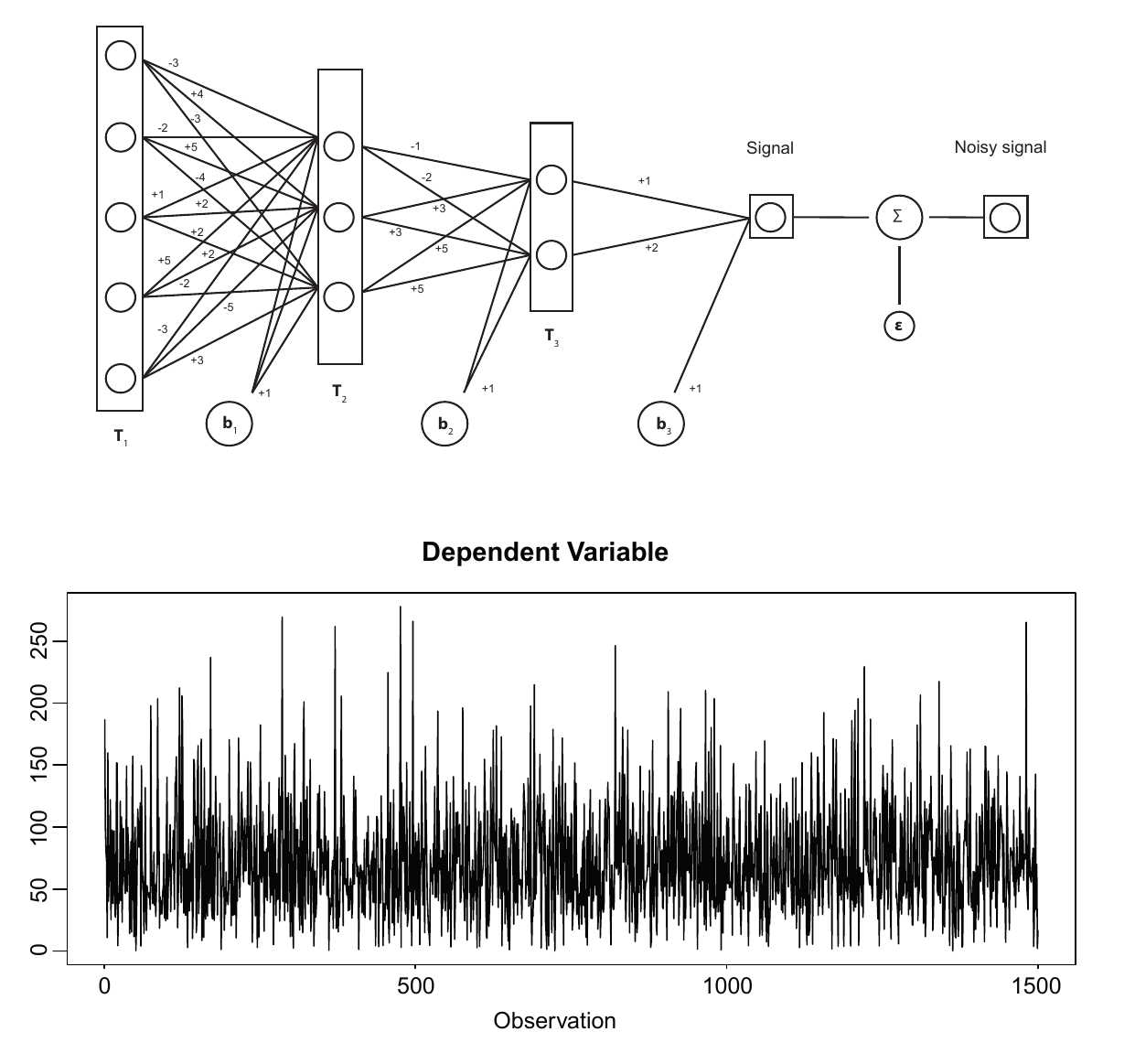}}
\caption[Data Generating Process]{Data Generating Process}
\label{visualDGP}
\end{figure}

\vspace{3mm} 
Following, a linear DGP that allows for interactions among the variables is also simulated. Also in this case $\mathbf{x} \in \mathbb{R}^{5}$, with $\mathbf{x}_{1} \sim\mathcal{N}(-4, 1)$, $\mathbf{x}_{2} \sim\mathcal{N}(1, 1)$, $\mathbf{x}_{3} \sim\mathcal{N}(1, 1)$, $\mathbf{x}_{4} \sim\mathcal{N}(1, 1)$, and $\mathbf{x}_{5} \sim\mathcal{N}(5, 1)$\footnote{The vector of means is generated from $U[-5, 5]$ and then rounded to the closest digit.}. The cross-correlation matrix is defined in \ref{corChole}. The analyzed DGP is\footnote{The interaction terms are introduced in order to have an unknown network structure. In fact, if no interactions are assumed, the true network structure is a shallow network with one hidden node.}

\begin{equation}
y = -8\mathbf{x}_{1} + 2\mathbf{x}_{2} + 2\mathbf{x}_{3} + 2\mathbf{x}_{4} + 7\mathbf{x}_{5} + 3\mathbf{x}_{1}\mathbf{x}_{2} - \mathbf{x}_{3}\mathbf{x}_{5} + 2\mathbf{x}_{1}\mathbf{x}_{4} + \epsilon
\end{equation} 

\vspace{3mm}
\noindent
The parameters chosen for the vector of coefficients are generated from a $U[-10, 10]$ and then rounded to the closest digit; the error term is $\epsilon \sim \mathcal{N}(0, 1)$ and it is uncorrelated with the input variables.

For both linear and nonlinear DGPs, a total of $1500$ observations are generated, $1200$ observations are used for the training set and $300$ for the test set. The datasets are normalized so that $\mathbf{x}$ has zero mean and unit variance. 

When fitting the neural networks, no optimal tuning of the neural network hyper-parameters and structure is conducted\footnote{For the correct choice of the network hyper-parameters, the analyst should ensure that the test set used for parameter tuning and for the aleatoric uncertainty computation is distinct - that is, a hold-out set should also be generated - otherwise, the consequential under-estimation of the aleatoric uncertainty could lead to narrower prediction intervals.}. The reasons for imposing the network hyper-parameters as opposed to fine-tuning them are: (I) it is ensured that the simulation results obtained are not dependent on fine-tuning; (II) it allows conducting a comparison of the empirical coverage rates across the three different methodologies analyzed, and (III) it allows analyzing the impact that different $p$s may have on the empirical coverage probabilities. 

When the nonlinear DGP is simulated, it is assumed that the neural network structure is known ($Z_{1} = 3$ and $Z_{2} = 2$). Conversely, when the linear DGP is analyzed - as the true network structure is unknown, and due to the simplicity of the DGP - a shallow network with $5$ hidden nodes is considered. When a nonlinear DGP is analyzed a $p = 0.995$ is applied (the true network structure is known and thus a low dropout rate is required); conversely, when a linear DGP is fitted - by imposing $p = [0.995, 0.990, 0.950, 0.900, 0.800]$ - it is possible to analyze the impact that different $p$s may have on the empirical coverage rates of the obtained prediction intervals. A sensible choice of the network parameters for the linear process is to use the Adam optimizer with learning rate $0.1$ and $10$ epochs; for the nonlinear process the Adam optimizer with learning rate $0.01$ and $80$ epochs. 

\subsection[Simulation Results]{Simulation Results}\label{normalSett}
Table \ref{experimentSimulation} reports the out-of-sample performance and the empirical coverages of the three procedures analyzed. When the nonlinear DGP is considered, one could notice that the three methodologies return - for the three different significance levels - prediction intervals with empirical coverage probabilities approximately equal to the theoretical ones. Focusing on the linear DGP, one could notice that the bootstrap approach returns prediction intervals with empirical coverages approximately equal to the significance level at which they are constructed; when the extra-neural network is considered, all prediction intervals - for the different $p$s considered - have an empirical coverage probability approximately equal to the nominal one; conversely, the MC dropout returns correct prediction intervals only for given values of $p$. 

As explained in the previous sections, the epistemic uncertainty in the MC dropout is captured exclusively by dropout at test time (and thus by the dropout rate $q = 1-p$). Conversely, when the extra-neural network is analyzed, the epistemic uncertainty depends not only on the dropout rate considered, but also on the random weight initialization used for fitting the $T$ sub-networks. As a result, the correct construction of the prediction intervals using the MC dropout approach requires to identify the optimal dropout rate as opposed to the extra-neural network algorithm proposed in the present paper. 

\bigskip

\begin{table}[H]
  \centering
  \captionsetup{singlelinecheck=false, justification = justified}
  \caption[Average test performance]{The table reports the out-of-sample mean average prediction error (MAPE) and mean squared prediction error (MSPE) for the analyzed procedures. EN$_{1}$ refers to the extra-neural network fitted for a nonlinear DGP, EN$_{2}$ for a linear DGP. MC$_{1}$ refers to the MC dropout for a nonlinear DGP, MC$_{2}$ for a linear DGP. Finally, BOOT$_{1}$ reports the results for the bootstrap approach a nonlinear DGP, BOOT$_{2}$ for a linear process.}
 \begin{adjustbox}{width=1\textwidth}
    \begin{tabular}{lr|r|r|rrrrr|rrrrr|r}
          & \multicolumn{2}{c}{\textbf{Nonlinear}} &       &       &       &       &       & \multicolumn{2}{c}{\textbf{Linear}} &       &       &       & \multicolumn{1}{r}{} &  \\
    \midrule
          & \multicolumn{1}{l|}{\textbf{EN}$_{1}$} & \multicolumn{1}{l|}{\textbf{MC}$_{1}$} & \multicolumn{1}{l|}{\textbf{BOOT}$_{1}$} &       &       & \multicolumn{1}{l}{\textbf{EN}$_{2}$} &       &       &       &       & \multicolumn{1}{l}{\textbf{MC}$_{2}$} &       &       & \multicolumn{1}{l}{\textbf{BOOT}$_{2}$} \\
    \midrule
    \textbf{p} & \multicolumn{1}{c|}{0.995} & \multicolumn{1}{c|}{0.995} & \multicolumn{1}{c|}{-} & 0.995 & 0.990  & 0.950  & 0.900   & 0.800   & 0.995 & 0.990  & 0.950  & 0.900   & 0.800   & \multicolumn{1}{c|}{-} \\
    \midrule
          T  = 30      & \multicolumn{1}{r}{} & \multicolumn{1}{r}{} & \multicolumn{1}{r}{} &       &       &       &       & \multicolumn{1}{r}{} &       &       &       &       & \multicolumn{1}{r}{} &  \\
    \midrule
    \textbf{MAPE} & 1.4979 & 3.5218 & 1.8476 & 1.0322 & 1.0493 & 1.1113 & 1.1196 & 1.2383 & 1.2993 & 1.3152 & 1.5834 & 1.6290 & 2.0478 & 1.0451 \\
    \textbf{MSPE} & 3.8232 & 19.8904 & 5.4190 & 1.7208 & 1.7544 & 2.0327 & 2.0315 & 2.6099 & 2.9998 & 2.9527 & 4.1508 & 4.0322 & 7.0050 & 1.7037 \\
    \textbf{Cov}$_{\mathbf{99}}$ & 0.01 & 0.01 & 0.01 & 0.01 & 0.01 & 0.01 & 0.01 & 0.01 & 0.02  & 0.01 & 0.00  & 0.01  & 0.00  & 0.01 \\
    \textbf{Cov}$_{\mathbf{95}}$ & 0.05 & 0.04  & 0.04  & 0.03  & 0.03  & 0.05  & 0.04 & 0.06  & 0.04  & 0.04 & 0.01  & 0.01  & 0.00  & 0.03 \\
    \textbf{Cov}$_{\mathbf{90}}$ & 0.07  & 0.08 & 0.06  & 0.09  & 0.09  & 0.08  & 0.10 & 0.08  & 0.07  & 0.07  & 0.02  & 0.02  & 0.00  & 0.09 \\
    \midrule
         T  = 50       & \multicolumn{1}{r}{} & \multicolumn{1}{r}{} & \multicolumn{1}{r}{} &       &       &       &       & \multicolumn{1}{r}{} &       &       &       &       & \multicolumn{1}{r}{} &  \\
    \midrule
    \textbf{MAPE} & 1.5068 & 3.5404 & 1.4480 & 1.0419 & 1.0808 & 1.0668 & 1.0940 & 1.2034 & 1.3044 & 1.3124 & 1.5337 & 1.5842 & 2.0583 & 1.0671 \\
    \textbf{MSPE} & 3.6592 & 20.0717 & 3.4133 & 1.7332 & 1.8930 & 1.7991 & 1.9732 & 2.4329 & 3.0670 & 2.8893 & 3.8274 & 3.9688 & 6.7150 & 1.8043 \\
    \textbf{Cov}$_{\mathbf{99}}$ & 0.01 & 0.01 & 0.01 & 0.01 & 0.01 & 0.01 & 0.01 & 0.01 & 0.01 & 0.01 & 0.00  & 0.01  & 0.00  & 0.01 \\
    \textbf{Cov}$_{\mathbf{95}}$ & 0.04 & 0.04 & 0.04 & 0.04  & 0.04  & 0.04  & 0.05 & 0.04  & 0.04  & 0.04  & 0.01  & 0.01  & 0.00  & 0.04 \\
    \textbf{Cov}$_{\mathbf{90}}$ & 0.08  & 0.08  & 0.09 & 0.09  & 0.09  & 0.09  & 0.09 & 0.10  & 0.08  & 0.08  & 0.01  & 0.02  & 0.00  & 0.08 \\
    \midrule
          T  = 70      & \multicolumn{1}{r}{} & \multicolumn{1}{r}{} & \multicolumn{1}{r}{} &       &       &       &       & \multicolumn{1}{r}{} &       &       &       &       & \multicolumn{1}{r}{} &  \\
    \midrule
    \textbf{MAPE} & 1.4756 & 3.5200 & 1.6426 & 1.0423 & 1.0467 & 1.1051 & 1.1611 & 1.1980 & 1.3026 & 1.3042 & 1.5277 & 1.5603 & 2.0060 & 1.0522 \\
    \textbf{MSPE} & 3.5096 & 20.1656 & 4.3616 & 1.7131 & 1.7315 & 1.9444 & 2.2202 & 2.4559 & 3.0330 & 2.9104 & 3.8339 & 3.8921 & 6.6385 & 1.7290 \\
    \textbf{Cov}$_{\mathbf{99}}$ & 0.01 & 0.01 & 0.01 & 0.01 & 0.01 & 0.02  & 0.01 & 0.01 & 0.01 & 0.00  & 0.00  & 0.01  & 0.00  & 0.01 \\
    \textbf{Cov}$_{\mathbf{95}}$ & 0.04 & 0.04 & 0.04 & 0.04 & 0.04 & 0.04  & 0.04  & 0.04 & 0.04 & 0.04  & 0.00  & 0.01  & 0.00  & 0.04 \\
    \textbf{Cov}$_{\mathbf{90}}$ & 0.09 & 0.08  & 0.08  & 0.10 & 0.10 & 0.09  & 0.09  & 0.10 & 0.07  & 0.08  & 0.01  & 0.02  & 0.00  & 0.08 \\
     \bottomrule
    \end{tabular}%
\end{adjustbox}
  \label{experimentSimulation}%
\end{table}%

\bigskip

The present research extends the results of Levasseur et al. (2017). Similarly, these authors state that the construction of prediction intervals with empirical coverage rates approximately equal to the theoretical ones - using the MC dropout approach - depends on the correct choice of the dropout rate. Consequentially, these authors suggest that the dropout rate should be tuned to return the correct prediction intervals. The theoretical analysis in Section \ref{prediction} coupled with the results in Table \ref{experimentSimulation} clearly show that the prediction intervals computed from the MC dropout rely significantly on the correct choice of the dropout rate. These results also suggest that choosing the dropout rate that maximizes the out-of-sample accuracy guarantees prediction intervals with the correct $\bar{\alpha}$ (the out-of-sample error is minimized for the $p$ that returns correct prediction intervals).

Although the results in Table \ref{experimentSimulation} show that all three procedures return prediction intervals with empirical coverage probabilities close to the theoretical ones for both linear and nonlinear DGPs, the performance of the extra-neural network approach is clearly superior in terms of coverage probabilities and also MAPE and MSPE errors. This is particularly the case for $\alpha = 0.10$. This outperformance is especially remarkable for the linear process for which we do not impose or know a priori the true structure of the network. Finally, focusing on the out-of-sample performance, one could notice that: (i) the out-of-sample errors decrease as $T$ increases, and (ii) for given dropout rates, the ensemble of neural networks outperforms the bootstrap approach. 

\subsection[Analysis of the correlation among learners]{Analysis of the correlation among learners} \label{sim_correlation}
An additional Monte Carlo simulation exercise is carried out to test if the assumption of independence among the single learners used in implementing the extra-neural network algorithm is correct, and thus if Equation \ref{predictionEnsemble1} is valid. In particular, while studying if $c_{i}$ in Equation \ref{refMSE} can be rightly assumed equal to zero, one should notice that it is important to respect the heteroscedasticity assumption used in computing the epistemic uncertainty, $\widehat{\sigma}^{2}_{\widehat{\bm{\omega}}}(\mathbf{x}_i)$. That is, the correlation among predictors should be computed for a given $\mathbf{x}_{i}$, returning $n$ correlations among the $T$ learners with $n$ being the length of the test sample. 

In order to do so, it is necessary to introduce variability in the ensemble predictions, while preserving the extra-neural network estimates. The strategy adopted in the present paper is to introduce uncertainty through bootstrap methods. The bootstrap procedure for $T$ learners is as follows: given $\{\mathbf{x}_{i}\}_{i=1}^{M}$ the set of covariates and $\{\mathbf{y}_{i}\}_{i=1}^{M}$ the target variable with $M$ the length of the training set and $\{\mathbf{x}_{i}\}_{i=1}^{n}$ the set of covariates in the test sample, each subnetwork $f_{t}$ with $t=1,\cdots, T$ is trained on $\{\mathbf{x}_{i},\mathbf{y}_{i}\}_{i=1}^{M}$ and the trained weights $\widehat{\bm{\omega}}^{(t)} = \{\mathbf{W}^{1, (t)}, \cdots, \mathbf{W}^{N, (t)}, b_{1}^{(t)}, \cdots, b_{N}^{(t)}\}$ are stored. Following, $\{\mathbf{x}_{i}\}_{i=1}^{n}$ is resampled with replacement $B$ times to obtain $B$ bootstrapped replica $\{\mathbf{x}_{i}^{\star, b}, \cdots, \mathbf{x}_{i}^{\star, B}\}_{i=1}^{n}$ with $b = 1, \cdots, B$. From the resampled datasets, $B$ out-of-sample predictions $\{\widehat{\mathbf{y}}_{i}^{\star, b}, \cdots, \widehat{\mathbf{y}_{i}}^{\star, B}\}_{i=1}^{n}$ are obtained for each $f_{t}$ for a fixed set of weights $\widehat{\bm{\omega}}^{(t)}$, with $t = 1, \cdots, T$. Starting from the bootstrapped predictions, the covariance among the $T$ learners is computed as follows: for each observation $\mathbf{x}_{i}$ with $i = 1, \cdots, n$, $B$ predictions from each subnetwork $f_{t}$ are stored. As a result, $n$ variance-covariance matrices ($T\times T$) are obtained and for each $\mathbf{x}_{i}$ it is possible to compute the average covariance between the $f_{t}$ subnetworks (thus satisfying Equation \ref{refMSE}). Below, the average covariance (average across the $n$ obtained covariances) is reported. 

The above bootstrap exercise is conducted imposing $B = 100$ and $T = 30$, when the underlying data generating process is nonlinear. To control for the validity of the bootstrap exercise, it is necessary to study if the bootstrap predictions of each $f_{t}$ learner are unbiased and, similarly, if the ensemble of the bootstrap predictions is an unbiased estimate of the out-of-sample target variable. For each of the $T = 30$ learners, the mean, the standard deviation, the kurtosis, the skewness, and the p-value from the Shapiro-Wilk test are collected. The results reported in Table \ref{correExperimentTable} show how the bootstrapped means are approximately equal to the mean prediction obtained on the test set $\{\mathbf{x}_{i}\}_{i=1}^{n}$, and that the null hypothesis of normality is rejected at $0.05$ significance level only five times\footnote{Two subnetworks are missing as the random fixed Bernoulli mask returned the null model.}. 

\begin{table}[H]
  \centering
  \captionsetup{singlelinecheck=false, justification = justified}
  \caption[Average test performance]{The table reports the summary statistics (mean, standard deviation, skewness, Kurtosis, and the p-value from the Shapiro-Wilk test for normality) for the bootstrap replica for each of the $T$ learner. Additionally, it also reports the true mean prediction (prior to resampling) for each subnetwork.}
 \begin{adjustbox}{width=1\textwidth}
    \begin{tabular}{rrrrrrr}
          & \multicolumn{1}{l}{\textbf{True Mean Pred.}} & \multicolumn{1}{l}{\textbf{Mean Boot.}} & \multicolumn{1}{l}{\textbf{Stand. Dev}} & \multicolumn{1}{l}{\textbf{Kurtosis}} & \multicolumn{1}{l}{\textbf{Skewness}} & \multicolumn{1}{l}{\textbf{Shaprio-Wilk}} \\
    \toprule
    \textbf{1} & 71.5608 & 71.7642 & 4.5904 & 3.2397 & 0.3521 & 0.0739 \\
    \textbf{2} & 71.9841 & 71.7823 & 4.4271 & 3.0619 & -0.0639 & 0.8900 \\
    \textbf{3} & 71.6901 & 71.8064 & 4.6287 & 2.6389 & 0.0480 & 0.6274 \\
    \textbf{4} & 71.4629 & 71.7165 & 4.7014 & 2.4896 & 0.1851 & 0.0229 \\
    \textbf{5} & 71.7296 & 71.9098 & 4.7610 & 2.8031 & -0.0253 & 0.6893 \\
    \textbf{6} & 72.1818 & 72.4281 & 4.2307 & 3.3075 & 0.0607 & 0.2826 \\
    \textbf{7} & 71.5742 & 71.2419 & 4.5291 & 3.0823 & 0.0541 & 0.7663 \\
    \textbf{8} & 71.6073 & 71.6189 & 4.4282 & 2.8779 & -0.0019 & 0.3129 \\
    \textbf{9} & 71.5663 & 71.4349 & 4.7200 & 2.5119 & 0.1792 & 0.0484 \\
    \textbf{10} & 71.4790 & 71.7649 & 4.9590 & 3.2489 & 0.3611 & 0.0143 \\
    \textbf{11} & 71.6780 & 71.9073 & 4.8519 & 3.1276 & 0.1247 & 0.9003 \\
    \textbf{12} & 72.0937 & 72.0630 & 4.4939 & 3.4614 & 0.5351 & 0.0007 \\
    \textbf{13} & 71.6108 & 71.8293 & 4.4362 & 2.6581 & -0.0052 & 0.2343 \\
    \textbf{14} & 71.5063 & 71.3657 & 4.2569 & 2.5846 & 0.2001 & 0.1142 \\
    \textbf{15} & 71.6539 & 71.5459 & 4.9013 & 3.6469 & 0.1230 & 0.0749 \\
    \textbf{16} & 71.5550 & 71.7832 & 4.7242 & 3.0683 & 0.1915 & 0.6314 \\
    \textbf{17} & 71.7348 & 71.5783 & 4.8394 & 2.7690 & 0.0973 & 0.5168 \\
    \textbf{18} & 71.7149 & 71.3114 & 4.4329 & 2.9711 & 0.0177 & 0.8693 \\
    \textbf{19} & 71.4578 & 71.6132 & 4.2455 & 2.4722 & 0.1635 & 0.0650 \\
    \textbf{20} & 71.6526 & 71.5065 & 4.6349 & 3.0424 & 0.1277 & 0.2733 \\
    \textbf{21} & 71.7326 & 72.2212 & 4.6082 & 3.0612 & -0.0401 & 0.8511 \\
    \textbf{22} & 71.5498 & 71.6016 & 4.9611 & 3.2911 & 0.3268 & 0.0115 \\
    \textbf{23} & 71.6915 & 71.7282 & 4.5501 & 3.0412 & 0.1021 & 0.4382 \\
    \textbf{24} & 71.4712 & 71.3337 & 4.8462 & 3.4061 & 0.1846 & 0.1460 \\
    \textbf{25} & 71.7813 & 72.1153 & 4.7040 & 3.2497 & -0.0248 & 0.8850 \\
    \textbf{26} & 71.5938 & 71.7936 & 4.6807 & 2.9165 & 0.1540 & 0.7936 \\
    \textbf{27} & 71.5279 & 71.3395 & 4.5650 & 2.5916 & -0.2339 & 0.0585 \\
    \textbf{28} & 71.5668 & 71.7543 & 4.6246 & 3.0035 & 0.2647 & 0.0993 \\
    \bottomrule
      \end{tabular}%
\end{adjustbox}
  \label{correExperimentTable}%
\end{table}%

The average covariance obtained with the procedure above is $0.0093$. Thus, these results show how the bootstrap samples from the extra-neural network predictions are independent and unbiased, with the average prediction over the test sample approximately normal. To further corroborate the assumption of independence, the average correlation (and absolute correlation) across the $T$ average bootstrapped predictions reported in Figure \ref{bootExtra} is computed. The average absolute correlation among the $T$ learners is $0.0498$, and the average correlation is $0.0124$. Finally, Figure \ref{bootExtra} reports the average bootstrap predictions for each of the $T$ learners considered, and the average out-of-sample observed target variable (in black). Also in this case, the average prediction is approximately equal to the average of the observed target variable, and we fail to reject the null hypothesis of normality from the Shapiro-Wilk test at $0.05$ significance level. Therefore, it is possible to conclude that also the ensemble of the bootstrap predictions is an unbiased estimate of the out-of-sample target variable. 

\vspace{3mm}
\begin{figure}[H]
\captionsetup{singlelinecheck = false, justification=justified} \centering
\framebox[\textwidth]{\includegraphics[page = 1, clip, trim= 0cm 18cm 0cm 0.5cm, width= 1\textwidth]{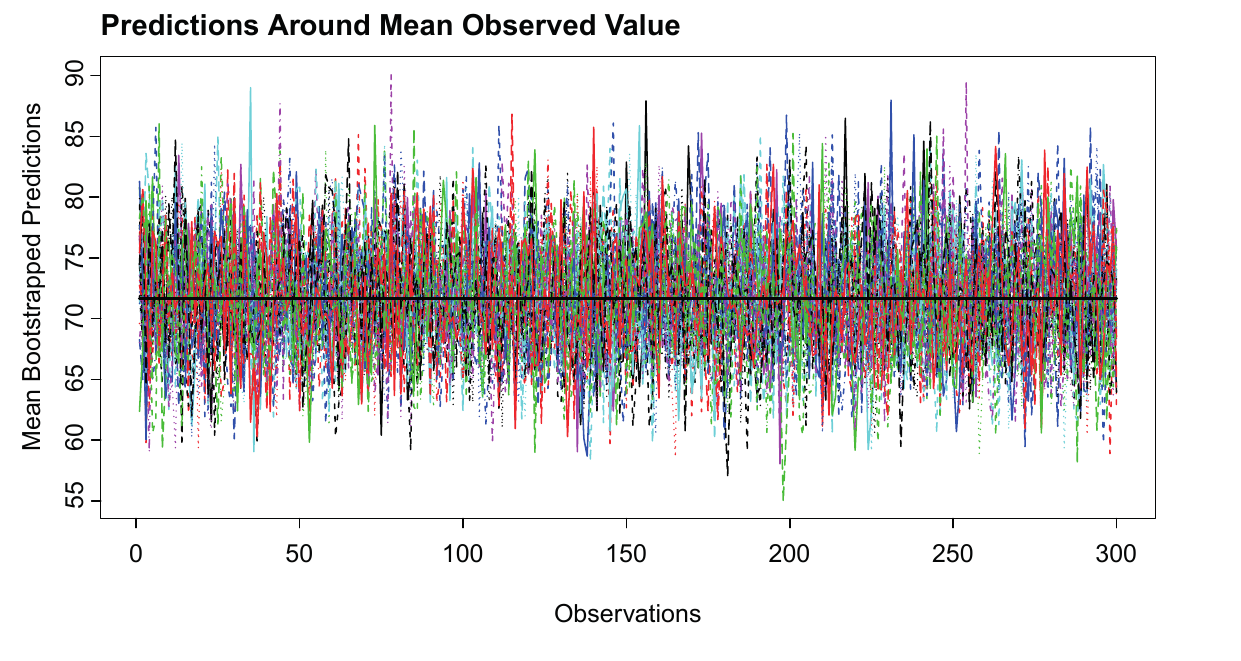}}
\caption[Mean bootstrapped predictions ($T=30$). In black the mean observed value is also reported.]{Mean bootstrapped predictions of the $T$ learners. In black the mean observed value is also reported.}
\label{bootExtra}
\end{figure}

To summarize, the simulation results prove that the predictions across the $T$ learners in the extra-neural network are independent and thus, that Equation \ref{predictionEnsemble1} is valid. Additionally, the results show that the proposed extra-neural network methodology not only returns correct prediction intervals but it also improves the forecast accuracy for both deep and shallow ensembles. Based on Equation \ref{refMSE}, one could also notice that the coverage probability of a prediction interval improves not only by correctly estimating the variance but also by providing more accurate pointwise predictions of the true observations. Therefore, the following section, by using the experimental settings of Hern\'{a}ndez-Lobato and Adams (2015), evaluates the out-of-sample accuracy in terms of root mean square prediction error (RMSPE) of the novel approach for real world datasets. 

\section[Empirical Analysis]{Empirical Analysis}\label{empiricalanalysislabel}
Hern\'{a}ndez-Lobato and Adams (2015) after proposing a novel scalable method for learning Bayesian neural networks  - called probabilistic backpropagation (PBP) - evaluate the performance of their novel methodology on real world datasets. The experimental settings used in their evaluation are widely adopted by the literature focusing on deep learning (see for example Gal and Ghahramani 2016a; and Lakshminarayanan et al., 2017) when evaluating novel algorithms. Therefore, using their experimental setup ensures comparability of the results with the variational inference method by Graves (2011), the probabilistic backpropagation of Hern\'{a}ndez-Lobato and Adams (2015), the MC dropout in Gal and Ghahramani (2016a), and the deep ensemble approach developed by Lakshminarayanan et al. (2017). 

The original experiment of Hern\'{a}ndez-Lobato and Adams (2015) evaluates the models not only in terms of RMSPE but also in terms of predictive log-likelihood (the latter being extremely relevant in Bayesian learning). Being the present paper focused on evaluating the accuracy of different procedures in constructing prediction intervals for regression tasks, only the former performance metrics will be considered. In fact, if the simulation - reported in the previous section - analyzes the correctness of the prediction intervals obtained from state-of-the-art methodologies designed not only for conditional mean but also variance estimation for both shallow and deep networks, it does not assess the performance of the extra-neural network approach for large datasets. Therefore, the present empirical application (focused on shallow structures in order to allow for cross-comparability) complements the results reported in Table \ref{experimentSimulation} by analyzing the RMSPE of the extra-neural network in large dimensional settings. The obtained RMSPEs (see also Equation \ref{refMSE}), by capturing both bias and variance of the predictions, provide an additional indication regarding the accurateness of the prediction intervals obtained from the extra-neural network algorithm. 

The experimental setup is as follows: $10$ datasets are analyzed. Each dataset is split into random training ($0.90$ of the observations) and test ($0.10$ of the observations) sets $20$ times and the average test set performance (RMSPE) and relative standard error are reported. As an exception, the protein and Year Prediction MSD datasets are split only $5$ and $1$ times into train and test sets. The datasets are normalized to guarantee that the regressors have zero mean and unit standard deviation. The same network architecture is considered: 1-hidden layer ReLu neural network with $Z_{1} = 50$ for the small datasets and $Z_{1} = 100$ for the larger protein and Year Prediction MSD datasets. Each neural network is trained for $40$ epochs. Following Gal and Ghahramani (2016a), we use a dropout rate of $0.05$, Adam optimizer and a batch size of $32$. We decide to use the same dropout rate as in Gal and Ghahramani (2016a) for comparability reasons. We refer to Gal and Ghahramani (2016a), Hern\'{a}ndez-Lobato and Adams (2015), and Lakshminarayanan et al. (2017) for additional details on the implementation of their algorithms. Lakshminarayanan et al. (2017) use $5$ networks in their ensemble, and Gal and Ghahramani (2016a) perform $10000$ stochastic forward passes\footnote{This is not directly reported by the authors and it is inferred from the code reported in their Github page (Gal and Ghahramani, 2016c).}. In order to allow for a fair comparison between the deep ensemble of Lakshminarayanan et al. (2017) and the novel algorithm proposed in the present paper, we will fit - at first - an extra-neural network with $5$ sub-networks; following, in order to compare the predictive performance of Algorithm \ref{alg1} with the MC dropout of Gal and Ghahramani (2016a), an extra-neural network with $70$ sub-networks will also be considered. 

\begin{table}[H]
  \centering
  \captionsetup{singlelinecheck=false, justification = justified}
  \caption[Average test performance]{The table reports the average test RMSPE and relative standard error (SE) for the variational inference method (VI) of Graves (2011); the probabilistic backpropagation (PBP) of Hern\'{a}ndez-Lobato and Adams (2015); the MC dropout of Gal and Ghahramani (2016a); and the deep ensemble proposed by Lakshminarayanan et al. (2017). Extra-net$_{1}$ uses $T = 70$, while Extra-net$_{2}$ uses $T = 5$. The number of observations used for the split is reported as $M+n$, and the dimension of the input as $d$. In bold the lowest average RMSPE is highlighted.}
 \begin{adjustbox}{width=1\textwidth}
    \begin{tabular}{lrrllllll}
    \textbf{Dataset} & \multicolumn{1}{l}{\textbf{(M+n)}} & \multicolumn{1}{l}{\textbf{d}} & \textbf{VI} & \textbf{PBP} & \textbf{MC-Dropout} & \textbf{Deep Ens.} & \textbf{Extra-net}$_{1}$ & \textbf{Extra-net}$_{2}$ \\
    \midrule
    Boston Housing  & 506   & 13    & 4.32$\pm$0.29 & 3.01$\pm$0.18 & 2.97$\pm$0.19 & 3.28$\pm$1.00 & \textbf{2.80}$\pm$\textbf{0.15} & 3.22$\pm$0.21 \\
    Concrete Strength & 1030  & 8     & 7.19$\pm$0.12 & 5.67$\pm$0.09 & 5.23$\pm$0.12 & 6.03$\pm$0.58 & 5.26$\pm$0.15 & \textbf{5.09}$\pm$\textbf{0.10} \\
    Energy Efficiency  & 768   & 8     & 2.65$\pm$0.08 & 1.80$\pm$0.05 & 1.66$\pm$0.04 & 2.09$\pm$0.29 & \textbf{0.59}$\pm$\textbf{0.01} & 0.72$\pm$0.02 \\
    Kin8nm & 8192  & 8     & 0.10$\pm$0.00 & 0.10+0.00 & 0.10$\pm$0.00 & 0.09$\pm$0.00 & \textbf{0.08}$\pm$\textbf{0.00} & \textbf{0.08}$\pm$\textbf{0.00} \\
    Naval Propulsion  & 11934 & 16    & 0.01$\pm$0.00 & 0.01$\pm$0.00 & 0.01$\pm$0.00 & \textbf{0.00}$\pm$\textbf{0.00} & 0.01$\pm$0.00 & 0.03$\pm$0.00 \\
    Power Plant  & 9568  & 4     & 4.33$\pm$0.04 & 4.12$\pm$0.03 & \textbf{4.02}$\pm$\textbf{0.04} & 4.11$\pm$0.17 & 4.12$\pm$0.05 & 4.24$\pm$0.04 \\
    Protein Structure & 45730 & 9     & 4.84$\pm$0.03 & 4.73$\pm$0.01 & 4.36$\pm$0.01 & 4.71$\pm$0.06 & \textbf{4.32}$\pm$\textbf{0.01} & 4.36$\pm$0.02 \\
    Wine Quality Red & 1599  & 11    & 0.65$\pm$0.01 & 0.64$\pm$0.01 & \textbf{0.62}$\pm$\textbf{0.01} & 0.64$\pm$0.04 & 0.63$\pm$0.01 & 0.64$\pm$0.01 \\
    Yacht Hydrodynamics & 308   & 6     & 6.89$\pm$0.67 & 1.02$\pm$0.05 & 1.11$\pm$0.09 & 1.58$\pm$0.48 & \textbf{0.72}$\pm$\textbf{0.06} & 0.97$\pm$0.06 \\
    Year Protection MSD & 515345 & 90    & 9.03$\pm$NA\tablefootnote{For the last dataset, it is not possible to compute the SE as only $1$ split is performed.} & 8.88$\pm$NA & 8.85$\pm$NA & 8.89$\pm$NA & \textbf{8.84}$\pm$\textbf{NA}\tablefootnote{If the predictions are rounded to the closest digit, or the floor operator is used, the obtained RMSE is 8.85.} & 8.97$\pm$NA \\
   \bottomrule
    \end{tabular}%
\end{adjustbox}
  \label{experimentEmpirical}%
\end{table}%

Table \ref{experimentEmpirical} reports the average RMSPE and relative standard errors; in bold are reported the lowest average RMSPEs. The authors - as opposed to what could be inferred from the related literature - indicate that it is not possible to ascertain the outperformance of one procedure over the competitors by relying solely on the average (over the resampled train and test sets) RMSPE; it is necessary to consider also the reported standard errors. Thus, the extra-network ($T = 70$) is shown to outperform the competing algorithms in four cases (excluding the Year Protection MSD dataset); both MC dropout and deep ensemble models are shown to outperform the other procedures in one case. When comparing the deep ensemble of Lakshminarayanan et al. (2017) and the extra-neural network ($T = 5$), the extra-neural network is shown to outperform five times, the deep ensemble three times\footnote{The deep ensemble proposed by Lakshminarayanan et al. (2017) is a novel algorithm that it is shown to consistently outperform classic bootstrap based approaches.}. As an additional robustness exercise, the extra-network algorithm is also implemented using deep structures: when the small datasets are considered, deep neural networks with $5$ hidden layers and $10$ hidden nodes each are trained; when the large datasets are analyzed, the depth of the subnetworks is $5$ with equal width of $20$ hidden nodes per layer. The results suggest that the outperformance of the extra-neural network approach over state-of-art deep learning algorithm is robust also for deep structures\footnote{The results --available upon request-- are not reported in Table \ref{experimentEmpirical} as they require changes in the experimental settings (network structure) of the original experiment of Hern\'{a}ndez-Lobato and Adams (2015).}.

\section[Conclusions]{Conclusions}\label{conclusionSection}

This article proposes a novel model based on an ensemble of deep neural networks. Our novel approach builds upon the work of Geurts et al. (2006) by extending the extremely randomized trees approach to ensembles of neural networks. The introduction of a Bernoulli mask allows for an additional randomization scheme in the prediction of the individual learners that ensures not only the correct construction of the prediction intervals, but also training the neural networks on the entire training set, better generalization performance due to randomized architecture structures, and accuracy gains due to an increase in the diversity among the members of the ensemble.The randomization across individual learners guarantees mutual independence across individual prediction models reducing the variance of the ensemble predictor by $1/T$, with $T$ the number of models comprising the ensemble prediction.

The performance of the proposed algorithms is assessed in a comprehensive Monte Carlo exercise. The simulation results show the excellent performance of the proposed approach in terms of mean square prediction error. Similarly, the empirical coverage probabilities obtained from the three competing ensemble methods assessed in this study (MC dropout, bootstrap approach, and extra-neural network) return coverage rates close to the nominal significance levels when tested using out-of-sample data. Nevertheless, the extra-neural network introduced in this paper is shown to outperform the competing models in most cases but more significantly for a $10\%$ significance level. Additionally, the simulation results also show the robustness of the extra-neural network approach to the choice of the dropout rate, as opposed to the MC dropout approach. In fact, in order to return correct prediction intervals with MC dropout , it is necessary to fine-tune the dropout rate that minimizes the out-of-sample error. This is not necessary for the extra-net approach.

These methods for prediction using ensembles of neural networks are further evaluated on real world datasets using the experimental settings of Hern\'{a}ndez-Lobato and Adams (2015). The results suggest that the extra-neural network approach outperforms state-of-the-art deep learning algorithms in terms of out-of-sample mean square prediction error.

\newpage

\section*{Appendix:  Random weight initialization}
Shallow and deep neural network are usually trained via the gradient descent (GD) algorithm that - being an iterative algorithm - requires an initial value for the parameter to be estimated.  Goodfellow et al. (2016) explain how - due to the difficulty in training neural networks (in particular DNNs) - training algorithms and thier convergence depend heavily on the choice of the initialization: different initial points can determine if the algorithm converges or not, if it converges to a global or local minimum, or the speed of convergence. Consequentially, it follows that different weight initialization will lead to different parameter ($\bm{\omega}$) estimates. More formally, consider Gaussian initialization and define $\{\mathbf{W}^{1}_{0},\ldots,\mathbf{W}^{N}_{0}\}$ as the weights generated at the beginning of the GD algorithm; by considering $e = 1, \cdots, E$ epochs, it is possible to define the GD update rule as: 

\begin{equation}\label{updateGD}
\mathbf{W}^{n}_{e} = \mathbf{W}^{n}_{e - 1} - \eta \nabla_{\mathbf{W}^{n}}L(\mathbf{W}^{n}_{e - 1}), \ \ \ \ \ n = 1, \cdots, N
\end{equation} 

\vspace{3mm} 
\noindent 
with $\eta$ being the learning rate and $\nabla_{\mathbf{W}^{n}}L(\mathbf{W}^{n}_{e - 1})$ being the partial gradient of the training loss $L(\mathbf{W}^{n}_{e - 1})$ with repsect to $\mathbf{W}^{n}$ defined as: 

\begin{equation}\label{updateGDsecond}
L(\mathbf{W}^{n}_{e - 1}) = \frac{1}{M}\sum_{i=1}^{M}L(f(\mathbf{x}_{i}; \widehat{\bm{\omega}});y_{i}), \ \ \ \ \ n = 1, \cdots, N
\end{equation} 
\noindent
with $M$ the number of observation in the train set.

From Equation \eqref{updateGD} and \eqref{updateGDsecond}, one could notice how the estimated $\{\mathbf{W}^{n}_{E}\}_{n=1}^{N}$ depends on $\{\mathbf{W}^{n}_{0}\}_{n=1}^{N}$, $\eta$, and the optimization algorithm implemented. Therefore, following the aforementioned literature and by assuming that both learning rate and optimization algorithm are equal across the different bootstrap realizations, the $\sigma^{2}_{\text{epistemic}}$ can be captured by allowing random weight initialization\footnote{The present analysis does not consider recent advances analyzing the relation between neural networks' dimensions ($Z_{\text{tot}}$) and weight initialization that ensures the presevation of the initialization properties during training. As an example, Zou et al. (2018) provide the condition under which Gaussian random initialization and (stochastic) GD produce a set ot iterated estimated weights that centers around $\{\mathbf{W}^{n}_{0}\}_{n=1}^{N}$ with a perturbation small enough to guarantee the global convergence of the algorithm, ultimately impacting on the approximation of the $\sigma^{2}_{\text{epistemic}}$ via random weight initialization.}.

\end{document}